\documentclass[final,5p,times,twocolumn,authoryear]{elsarticle}

\usepackage{amssymb}
\usepackage{colortbl}
\usepackage{bm}
\usepackage{amsmath}
\usepackage[caption=false]{subfig}
\usepackage{textcomp}

\usepackage{multirow}
\usepackage{booktabs}
\usepackage{cleveref}
\Crefname{figure}{Fig.}{Figs.}
\Crefname{appendix}{}{}
\Crefname{equation}{Eq.}{Eqs.}
\graphicspath{{Figs/}}

\usepackage{xspace}
\def\ie{\emph{i.e}.\null\xspace}

\usepackage{amsthm}
\theoremstyle{definition}
\newtheorem{theorem}{Theorem}

\theoremstyle{plain}
\newtheorem{lemma}[theorem]{Lemma}

\definecolor{mypink}{rgb}{.99,.91,.95}

\journal{Image and Vision Computing}

\begin{document}

\begin{frontmatter}



\title{CoDeGAN: Contrastive Disentanglement for Generative Adversarial Network}
% \tnoteref{label1}}
\author[uestc]{Jiangwei Zhao}
\author[uestc]{Zejia Liu}
\author[uestc]{Xiaohan Guo}
\author[uestc]{Lili Pan\corref{cor1}}
\cortext[cor1]{Corresponding author.}
\affiliation[uestc]{organization={
            School of Information and Communication Engineering, University of Electronic Science and Technology of China},%Department and Organization
            city={Chengdu},
            postcode={611731}, 
            country={China}}

% \author{}

% \affiliation{organization={},%Department and Organization

\begin{abstract}

Disentanglement, a critical concern in interpretable machine learning, has also garnered significant attention from the computer vision community. Many existing GAN-based class disentanglement (unsupervised) approaches, such as InfoGAN and its variants, primarily aim to maximize the mutual information (MI) between the generated image and its latent codes. However, this focus may lead to a tendency for the network to generate highly similar images when presented with the same latent class factor, potentially resulting in mode collapse or mode dropping. To alleviate this problem, we propose \texttt{CoDeGAN} (Contrastive Disentanglement for Generative Adversarial Networks), where we relax similarity constraints for disentanglement from the image domain to the feature domain. This modification not only enhances the stability of GAN training but also improves their disentangling capabilities. Moreover, we integrate self-supervised pre-training into \texttt{CoDeGAN} to learn semantic representations, significantly facilitating unsupervised disentanglement. Extensive experimental results demonstrate the superiority of our method over state-of-the-art approaches across multiple benchmarks.
The code is available at https://github.com/learninginvision/CoDeGAN.
\end{abstract}


% \begin{highlights}
% \end{highlights}


\end{frontmatter}


\section{Introduction}
\label{sec:intro}

Considerable efforts have been dedicated to enhancing interpretable and controllable generation~\citep{doshi-velez2017towards,lakkaraju2016interpretable,rudin2019stop,zhang2018interpretable} through constructing controllable deep generative models. Notably, several models have been successfully proposed, including InfoGAN~\citep{chen2016infogan} and its variants~\citep{lin2020infogan,ojha2020elastic,mukherjee2019clustergan}, Factor-VAE~\citep{kim2018disentangling}, and $\beta$-VAE~\citep{higgins2016beta}. Within these endeavors, representation disentanglement~\citep{gabbay2020demystifying,liu2020oogan,locatello2020disentangling,shu2020weakly} has garnered significant attention. 

Representation disentanglement refers to the problem of learning a representation that can separate the distinct and informative factors of variations in data. 
In recent years, the majority of literature concerning disentanglement~\citep{kingma2014auto,higgins2016beta,gabbay2020demystifying,liu2020oogan,locatello2020disentangling,shu2020weakly,kim2018disentangling, chen2016infogan, kim2021contrastive,hwang2021stein} has concentrated on Bayesian generative models like VAE~\citep{kingma2014auto}, $\beta$-VAE~\citep{higgins2016beta}, and Factor-VAE~\citep{kim2018disentangling}.
These models allow for the learning of representation through latent variable models with some structure prior. 
In contrast, scant attention~\citep{chen2016infogan, kim2021contrastive,hwang2021stein} has been devoted to GAN-based disentanglement, although GANs have a remarkable ability in generation compared to VAE.
We suspect this discrepancy arises from the inherent difficulty of posterior inference in GANs, which has posed a significant challenge to GAN-based disentanglement~\citep{mohamed2016learning, pan2020LDAGAN}.

%Besides, disentanglement may potentially induce mode collapse/dropping in training, as it expects the same valued factor to produce images with similar features~\citep{chen2016infogan,lin2020infogan,mukherjee2019clustergan}.
%Then, a natural question arises, without a probabilistic framework in GANs, how to learn a disentangled representation.

Maximizing mutual information (MI)~\citep{chen2016infogan} is one of the few choices in the GANs family that can produce disentangled representations. 
Thus, InfoGAN and its variants have been widely used in multi-factor (\ie, background, object, and texture) disentanglement to facilitate conditional image generation \citep{kim2021contrastive,benny2020onegan,li2020mixnmatch}.
Despite their wide usage in real-world applications, no previous study shows they may inherently incur the instability of GAN's training, and deteriorate disentanglement performance.
In InfoGAN, MI is defined as the mutual information between the generated image and the latent code.
Specifically, if $c$ denotes the latent code and $\mathbf{z}$ denotes in-compressible noise, the MI term is defined as:
\begin{equation}
	I\left(G(\mathbf{z},c); c\right)=H\left[G\left(\mathbf{z},c\right)\right]-H\left[G\left(\mathbf{z},c\right)|c\right],
\label{eq:MI}
\end{equation}
\noindent where $H\left(\cdot\right)$ denotes entropy and $G\left(\mathbf{z}, c\right)$ denotes the synthesised image. Maximizing \Cref{eq:MI} may significantly limit diversity, especially when $c$ is discrete, as it maximizes the reduction in uncertainty about $G\left(\mathbf{z}, c\right)$ under being told the value of $c$. 
This may incur the instability of GAN's training, potentially leading to mode collapse or mode dropping.
Although in InfoGAN, the actual loss is the variational lower bound on the mutual information, it enforces that images generated with the same latent code $c$ can be encoded to $c$.
Such a loss would also reduce generative diversity dramatically, hindering generative distribution from getting close to real distribution.
To alleviate this problem, especially in discrete factor disentanglement (\,\ie , class disentanglement), we propose contrastive disentanglement in GANs, where we relax the similarity constraint to the feature domain, rather than the input image domain.
%Furthermore, we provide a comprehensive theoretical analysis of why our method is prone to improve GANs equilibrium.
%Our work is the \emph{first} study to investigate GAN-based disentanglement from the perspective of GANs Equilibrium.

\begin{figure*}[t]
	\centering
	\includegraphics[width=0.90\textwidth]{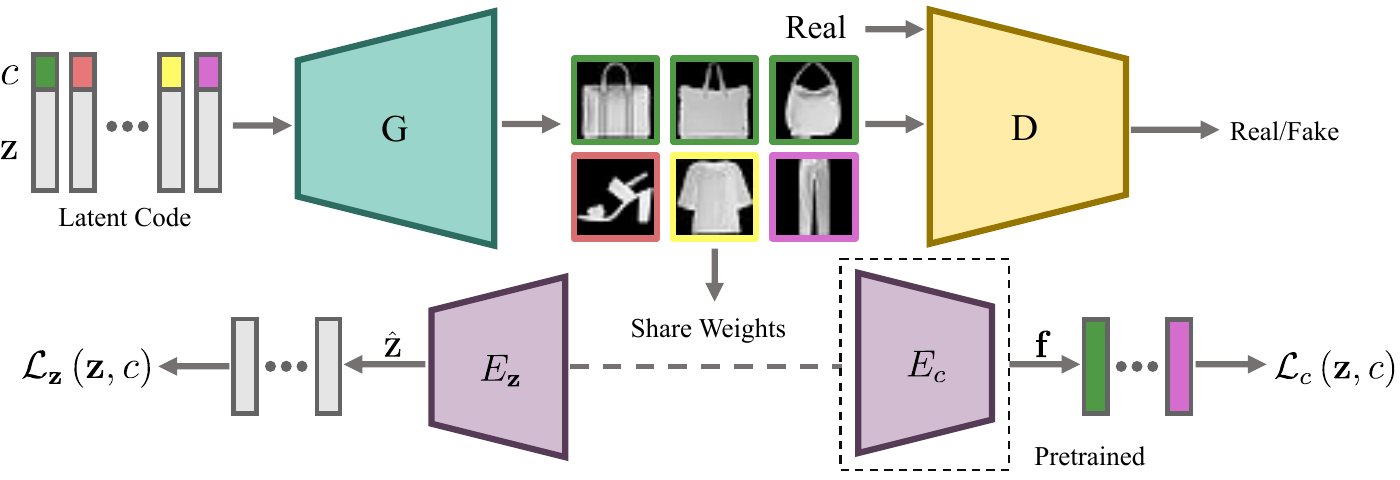}
	\caption{\textbf{Contrastive disentanglement framework.} The input of generator G consists of two parts: (\romannumeral1) $c\sim mul\left(\bm{\pi}\right)$, which controls the detailed class and (\romannumeral2) $\mathbf{z}\sim\mathcal{N}\left(\mathbf{0}, \sigma^2\mathbf{I}\right)$, which corresponds to intra-class variation. The representation $\mathbf{f}$ encoded by $E_c$ is regularized by contrastive loss $\mathcal{L}_c$ for disentanglement, while the representation $\hat{\mathbf{z}}$ encoded by $E_{\mathbf{z}}$ is regularized by reconstruction loss $\mathcal{L}_{\mathbf{z}}$ for preserving intra-class variation. $E_c$ and $E_{\mathbf{z}}$ could share weights partly. The definitions of positive and negative pairs in $\mathcal{L}_c$ are images within the same class or not.}
	\label{fig:Struc}
\end{figure*}

Another problem in maximizing MI-based GAN disentanglement approaches is that they learn image generation and latent factor disentanglement jointly.
Without any guidance, they have difficulty in learning univocal latent factors. 
As of now, such a problem almost remains unexplored.
Inspired by the rapid development of self-supervised pre-training methods~\citep{hinton2020SIMCLR,2019Momentum, wu2018unsupervised,jaiswal2021survey} in recent years, we recognize that pre-training may facilitate learning semantically meaningful representations for unsupervised disentanglement.
To this end, we integrate self-supervised pre-training into our contrastive disentanglement framework, striving to enforce the encoder to learn semantic representation in disentanglement.
As illustrated in \Cref{fig:Struc}, the encoder for extracting class-related features can be pre-trained.
% Besides, we experiment with using extremely limited labeled samples to further facilitate disentanglement. 

To sum up, this work shows two inherent problems in GAN-based disentanglement, and to alleviate these two problems, we propose \textbf{Co}ntrastive \textbf{D}is\textbf{e}ntanglement for \textbf{G}enerative \textbf{A}dversarial \textbf{N}etworks (\texttt{CoDeGAN}).
On the one hand, the proposed model, by relaxing the similarity constraint, can improve performance in both disentanglement accuracy and generative quality. 
On the other hand, \texttt{CoDeGAN} can naturally integrate self-supervised pre-training into GAN-based disentanglement, further improving disentanglement accuracy.
As shown in \Cref{fig:ACCFID}, we improve GAN's training stability, resulting in significantly improved Fr\'{e}chet Inception Distance (FID) and ACC.
On the CIFAR-10 dataset, we gain a stunning 19\% absolute improvement over InfoGAN and 16\% absolute improvement over the previous SOTA methods.
Not only that, we achieve the SOTA performance on other datasets.
Our findings will offer some fresh insights into the field of GAN-based disentanglement.
%------------------------------------------------------------------------
\section{Related Work}
% \label{sec:formatting}
\label{RelatedWorks}
Our work relates to several research fields.
We describe prior work for these broad fields, including disentanglement approaches based on variational autoencoders (VAEs) and generative adversarial networks (GANs), as well as self-supervised pre-training.

\subsection{VAE-based Disentanglement.} VAE-based disentanglement methods~\citep{chen2018isolating,higgins2016beta,kim2018disentangling} attempt to learn disentangled representations by encouraging the posterior distribution of latent code to be close to certain prior distributions. 
$\mathbf{\beta}$-VAE~\citep{higgins2016beta} uses a larger weight on the KL divergence between the variational posterior and the prior to fulfill disentanglement. 
$\mathbf{\beta}$-TCVAE~\citep{chen2018isolating} estimates the total correlation (TC), a measure of the statistical dependence or redundancy among the latent variables, by minibatch-weighted sampling, and places more emphasis on minimizing the TC.
This encourages the latent variables to become more independent and disentangled.
Factor-VAE~\citep{kim2018disentangling} utilizes a discriminator network to predict latent code to estimate the TC, avoiding the computational complexity of directly calculating high-dimensional TC.
Recently, the work~\citep{locatello2020disentangling} has proposed to use few labels to disentangle factors of variation under the VAE framework.
These methods have been shown to be capable of disentangling \emph{continuous} factors with some success; however, they are not well-suited to capture \emph{discrete} factors.

In contrast, our research focuses on disentangling \emph{discrete} factors, a direction that has received relatively little attention in VAE-based disentanglement approaches.

%------------------------------------------------------------------------
\subsection{GAN-based Disentanglement.}
GANs-based disentanglement methods~\citep{esser2020a,chen2016infogan,lin2020infogan,mukherjee2019clustergan,shen2020interpreting} can be categorized into two classes: unsupervised and supervised.
In this work, we mainly focus on unsupervised disentanglement.
InfoGAN~\citep{chen2016infogan}, as one typical unsupervised method, maximizes the mutual information between an image and its latent codes. 
InfoGAN-CR~\citep{lin2020infogan}, compared to InfoGAN, adds a contrastive regularizer (CR) to encourage paired images with shared factor to be similar. 
Although InfoGAN-CR uses the concept of contrastive regularization, it does not include the contrastive loss into the model.
Elastic-InfoGAN~\citep{ojha2020elastic} introduces contrastive loss as an auxiliary loss of InfoGAN, and forces the encoder to learn features focusing on image classes to disentangle class-imbalanced data.
ClusterGAN~\citep {mukherjee2019clustergan} trains the GANs and the inverse-mapping network under a clustering-specific loss. 

Although the idea of contrastive learning has been explored in some of the above disentanglement works~\citep{ren2021generative,kim2021contrastive,hwang2021stein}, our work is the \emph{first} study that explores removing the MI term and relies only on the contrastive loss for discrete factor disentanglement.

Except for unsupervised discrete factor disentanglement, there exist some other studies for unsupervised continuous factor disentanglement. 
InterFaceGAN~\citep{esser2020a} proposed an invertible interpretation network to disentangle the representations of semantic concepts.
% \Pan{ues the name of the method, but not the name of authors}.
StyleGAN~\citep{karras2019style} has achieved significant success in image generation, and the decoupling of its latent space is an important direction for disentanglement.
Analyzing changes in generated images by shifting latent codes enables the identification of relationships between latent codes and attributes of the generated images.
GANSpace~\citep{harkonen2020ganspace} uses principal component analysis to decompose the latent space so as to discover image attribute editing directions in StyleGAN.
% identifies important latent directions based on Principal Component Analysis (PCA) applied either in latent space or feature space, and creates a StyleGAN-like manner to control the image synthesis. 
SeFa~\citep{shen2021closed} proposes a closed-form factorization algorithm for latent semantic factor discovery.
%It extends the disentanglement problem to finding the eigenvectors associated with the k largest eigenvalues of the matrix product and the eigenvector is the disentanglement direction.
SRE~\citep{kappiyath2022self} proposes a scale ranking estimator, which distinguishes and captures the most important factors of variation by ranking the magnitude of variation of the latent code in the generated image along each direction.
DisCo~\citep{ren2021generative} proposed a model-agnostic method to disentangle continuous factors, where a navigator is trained with the contrastive loss to provide disentanglement directions.
It distinctly differs from our work as it is for continuous factor disentanglement.

In contrast to previous unsupervised disentanglement works, our method has three obvious  differences: 
(\romannumeral1) it has no MI term,
(\romannumeral2) it only relies on the contrastive loss for discrete factor disentanglement, and
(\romannumeral3) it leverages pre-training to extract semantically meaningful features to facilitate disentanglement.

%Recently, Locatello~\etal \cite{locatello2019challenging} has demonstrated that unsupervised disentanglement learning without inductive biases is theoretically impossible and
%that existing inductive biases and unsupervised methods do not allow to consistently learn disentangled representations.
%Some work~\citep{locatello2020disentangling} have pointed out that  using a limited amount of supervision is effective in VAE-based disentanglement methods.
%Similarly, the work~\citep{nie2020semi}, based on StyleGAN, designs unsupervised code reconstruction loss to introduce little supervision, and achieves better disentanglement performance.
%In this paper, we will also investigate GAN-based disentanglement with little supervision.
	
%------------------------------------------------------------------------

\subsection{Self-supervised pre-training.} 
Self-supervised pre-training is a form of unsupervised
training that captures the intrinsic patterns and properties of data without using human-provided labels to learn discriminative representations for downstream tasks~\citep{sohn2016improved,hinton2020SIMCLR,2019Momentum, wu2018unsupervised,2020Supervised,oord2018representation,wang2021dense,caron2020unsupervised,he2022masked}.
In computer vision, SimCLR first establishes a general framework for self-supervised pre-training using a contrastive scheme~\citep{hinton2020SIMCLR}. 
Momentum Contrast (MoCo)~\citep{2019Momentum, wu2018unsupervised} further demonstrates the effect of increasing negative pairs on promoting contrastive learning (CL) performance. 
Later, SimSiam~\citep{chen2021Siamese} has shown that stop-gradient operation plays an essential role in preventing collapse solutions, and even not using negative sample pairs, large batches and momentum encoders can learn meaningful representations.
Meanwhile, some theoretical papers~\citep{arora2019theoretical,oord2018representation,tschannen2019mutual,wang2020understanding,wu2020mutual,hjelm2018learning} discuss the relationship between contrastive self-supervised learning and the InfoMax principle, and interpret the success of CL from this perspective. 
%Furthermore, some new data augmentation methods~\citep{misra2020self, caron2020unsupervised, tian2020makes}, benefiting from reducing MI between views, gain performance improvement on various downstream tasks.
With the success of vision transformer (ViT), DINO~\citep{caron2021emerging}, which trains a ViT in a self-supervised manner, further bridges the gap between self-supervised pre-training and ViT. 
MAE~\citep{he2022masked} adopts masked image modeling (MIM) in self-supervised pre-training, which uses an asymmetric encoder-decoder to reconstruct the masked input image patches, and achieves improved performance in downstream tasks.
CLIP\citep{radford2021learning} pre-trains one textual encoder and one image encoder for text-image alignment on a large-scale text-image dataset.
BLIP\citep{li2022blip} is a new VLP framework that enables a wider range of downstream tasks. 
It introduces a multimodal mixture of Encoder-Decoder structure for unified vision-language understanding and generation, which can effectively perform multi-task pre-training and transfer learning.

In this work, we hope to explore how self-supervised pre-training can facilitate representation disentanglement.

%-------------------------------------------------------------------------
\section{Contrasitve Disentanglement in GANs}

%-------------------------------------------------------------------------
\subsection{Contrastive Disentanglement Framework}
Generative adversarial networks (GANs), as a fundamental deep generative model, directly learn a generator $G$ mapping from input latent variables $\mathbf{z}$ to image, and a discriminator $D$ to discriminate real and fake images.
Specifically, it trains $D$ to maximize the probability of correct discriminating between real and generative images and simultaneously trains $G$ to maximize the probability of the generated image being recognized as real ones.
The objective function is typically formulated as follows:
\begin{align}
	\min_{G}
	\max_{D}
	\mathbb{E}_{\mathbf{x} \sim p_{data}\left(\mathbf{x}\right)}
	&\left[ \log D\left( \mathbf{x}\right) \right] \nonumber\\
	&+\mathbb{E}_{\mathbf{z} \sim p\left(\mathbf{z}\right)}
	\left[ \log \left( 1-D( G\left( \mathbf{z}\right)\right) \right].
	\label{eq:GAN}
\end{align}
Here, $D\left(\mathbf{x}\right)$ represents the probability of $\mathbf{x}$ being real, while $D\left(G\left(\mathbf{z}\right)\right)$ pertains to the generated image $G\left(\mathbf{z}\right)$.

The task of disentanglement in GANs presents a challenge, as it is difficult to formulate a latent variable model with a data structure prior. In order to overcome this obstacle, we propose to disentangle the class variation of data by comparing the features of generated images. Our approach is based on the idea that the same factor will create similar image features that are associated with the factor.

To disentangle the class variation, which is the commonest in visual data, we define the input latent code consists of two parts: (\romannumeral1) $c\sim mul\left(\bm{\pi}\right)$, which controls the detailed class and (\romannumeral2) $\mathbf{z}\sim\mathcal{N}\left(\mathbf{0}, \sigma^2\mathbf{I}\right)$, which corresponds to intra-class variation.
Then, for each generated image $G\left(\mathbf{z}, c\right)$, given the discriminator $D$, the loss for generator now change to be:
\begin{equation}
\mathcal{L}_{GAN}=\mathbb{E}_{\mathbf{z} ,c \sim p\left(\mathbf{z}, c\right)}
	\left[ \log \left( 1-D( G\left( \mathbf{z}, c\right)\right) \right],
\end{equation}
where $p\left(\mathbf{z}, c\right)$ can be rewritten as $p\left(\mathbf{z}\right)p\left(c\right)$ as $\mathbf{z}$ and $c$ are independent variables.

Inspired by the intuition that the same latent class variable should produce images in the same class, we use an encoder to extract representation $\mathbf{f}$, \ie, $\mathbf{f}=E_{c}\left(G\left(\mathbf{z},c\right)\right)$ and construct contrastive loss $\mathcal{L}_c$ for feature $\mathbf{f}$.
We hope the same $c$, even combined with different $\mathbf{z}$, produces similar features $\mathbf{f}$, and vice versa.
Let $p_{pos}(\cdot,\cdot)$ be the positive distribution about $\left(\mathbf{z}, c\right)$ over $\mathbb{R}^{D+1}\times\mathbb{R}^{D+1}$, $p\left(\mathbf{z}, \text{neg}\left(c\right)\right)$ be the distribution of negative samples.
Here, $\text{neg}\left(c\right)\sim mul\left(\bm{\pi}\right), \pi_k=1/K, \text{neg}\left(c\right) \neq c$.
By comparing positive and negative pairs, we formulate $\mathcal{L}_c\left(\mathbf{z},c\right)$ as:
\begin{equation}
% \small{
	\displaystyle \mathop{\mathbb{E}}
	_{\substack{\left(\mathbf{z}, c, \mathbf{z}^+, c^+\right)\sim p_{\text{pos}}\\ \left(\mathbf{z}_i,c^-_i\right)\stackrel{i.i.d.}{\sim} p\left(\mathbf{z}, \text{neg}\left(c\right)\right)}} 
	\left[-\log \frac{ e^{\left(\mathbf{f}^T\mathbf{f}^+/\tau\right)}}{e^{\left(\mathbf{f}^T\mathbf{f}^+/\tau\right)} +
		\sum_i e^{\left(\mathbf{f}^T\mathbf{f}_i^-/\tau\right)}}\right],
	\label{eq:CL}
\end{equation}
where $\tau > 0$ is a scalar temperature hyper-parameter.
Here, $\mathbf{f}^+$ is the feature from positive samples, \ie, $E_c\left(G\left(\mathbf{z}^+,c^+\right)\right)$, $c^+ = c$, $\mathbf{z}^+$ is a sampling point from $p\left(\mathbf{z}\right)$ and $\mathbf{z}\neq \mathbf{z}^+$.
Each $\left(\mathbf{z}, c\right)$ has one positive pair and $K-1$ negative pairs. 
Similarly, $\mathbf{f}_i^-$ is the feature from negative samples, \ie, $E_c\left(G\left(\mathbf{z}_i,c_i^-\right)\right)$, $c_i^- \neq c$.
\Cref{eq:CL} encourages the generated images with the same latent code $c$ to have similar features $\mathbf{f}$, and the images with different latent code $c$ have unlike $\mathbf{f}$, as shown in \Cref{fig:Struc}.
Through such a contrastive loss, we can formulate the correlation between the class of generative images and latent code $c$.

Except enforcing the features associated with the class to be similar, we also expect the encoded representation $\hat{\mathbf{z}}= E_{\mathbf{z}}\left(G\left(\mathbf{z}, c\right)\right)$ to preserve intra-class variation~\citep{mukherjee2019clustergan} of data.
Here, $E_{\mathbf{z}}$ is the encoder responsible for extracting class-unrelated content.
Hence, we formulate a reconstruction loss $\mathcal{L}_\mathbf{z}\left(\mathbf{z},c\right)$ between the latent variables $\mathbf{z}$ and representation $\hat{\mathbf{z}}$,
\begin{equation}
	\mathcal{L}_\mathbf{z} \left(\mathbf{z},c\right)= \displaystyle \mathop{\mathbb{E}}_{\left(\mathbf{z},c\right)\sim p\left(\mathbf{z},c\right)}
	\|\mathbf{z}-\hat{\mathbf{z}}\|_2^2.
 \label{eq:Recon}
\end{equation}
The smaller $\mathcal{L}_\mathbf{z}$ is, the better the noise $\mathbf{z}$ has been preserved.
The reconstruction loss $\mathcal{L}_{\mathbf{z}}$ enforces the representations to be separate, obeying underlying Gaussian distribution, and thus has the potential to prevent mode collapse/dropping.

Finally, the objective function $\mathcal{L}_G $ takes account into the image reality loss $\mathcal{L}_{GAN}$, the contrastive loss $\mathcal{L}_c$ and the reconstruction loss $\mathcal{L}_{\mathbf{z}}$: 
\begin{equation}
	\begin{split}
		\mathcal{L}_G =&\mathbb{E}_{\mathbf{z} ,c \sim p\left(\mathbf{z}, c\right)}
		\left[ \log \left( 1-D( G\left( \mathbf{z}, c\right)\right) ) \right] \\ 
		&+ \beta_1 \mathcal{L}_c\left(\mathbf{z},c\right) 
		+ \beta_2 \mathcal{L}_{\mathbf{z}}\left(\mathbf{z},c\right),
	\end{split}
\end{equation}
where $\beta_1$ and $\beta_2$ are two trade-offs.
To constrain the flexibility, we prefer to learn $G$ and $E=\left\{E_c,E_{\mathbf{z}}\right\}$ separately. 
The learning procedure of \texttt{CoDeGAN} is a three-step alternative optimization procedure: (\romannumeral1) update discriminator $D$, (\romannumeral2) update generator $G$, and (\romannumeral3) update encoder $E$.

We acknowledge the lower bound on mutual information $I\left(G(\mathbf{z},c); c\right)$ in InfoGAN is indeed a cross-entropy loss between the real class distribution and posterior distribution $Q\left(c|G\left(\mathbf{z},c\right)\right)$. 
Such a loss enforces similarity among generated images within an exceedingly low-dimensional feature space, potentially diminishing generative diversity.
As we discussed previously, diminishing the generative diversity may incur the instability of GAN's training, resulting in deteriorated disentanglement accuracy.

The reconstruction loss in \texttt{CoDeGAN}, as defined in \Cref{eq:Recon}, bears some resemblance to the previously mentioned cross-entropy loss, as both aim to ensure proximity between the encoded features and the input latent variables. However, owing to the significantly higher dimensionality of the intra-class variables $\mathbf{z}$ and their distinctiveness across each sampling, the reconstruction loss imposed on $\mathbf{z}$ does not diminish diversity but rather encourages it.
%TODO:?
%-------------------------------------------------------------------------

\subsection{Theoretical Insights}
%-------------------------------------------------------------------------
\label{sec:ConDisA}
To analyze the contrastive disentanglement loss, we analyze the lower bound of $\mathcal{L}_c\left(\mathbf{z},c\right)$. 

\begin{lemma}
Minimizing the contrastive loss $\mathcal{L}_c$ is equivalent to maximizing the mutual information between generative representations $\mathbf{f}$ and $\mathbf{f}^+$, \ie, $E_c\left(G\left(\mathbf{z},c\right)\right)$ and $E_c\left(G\left(\mathbf{z}^+,c^+\right)\right)$.
\begin{equation}
\label{eq:bound}
    \begin{split}
		\displaystyle \mathop{\mathbb{E}}
		_{\substack{\left(\mathbf{z}, c, \mathbf{z}^+, c^+\right)\sim p_{\text{pos}}\\
		\left(\mathbf{z}_i,c^-_i\right)\stackrel{i.i.d.}{\sim} p\left(\mathbf{z}, \text{neg}\left(c\right)\right)}}  
		&\left[-\log \frac{ e^{\left(\mathbf{f}^T\mathbf{f}^+/\tau\right)}}{e^{\left(\mathbf{f}^T\mathbf{f}^+/\tau\right)} +
			\sum_i e^{\left(\mathbf{f}^T\mathbf{f}_i^-/\tau\right)}}\right]\\ 
		&\geq\log\left(K\right)-I\left(\mathbf{f}, \mathbf{f}^+\right).
		\end{split}
		% }
\end{equation}
\end{lemma}
\noindent This lower bound indicates that the contrastive disentanglement loss is dominated by the mutual information between $\mathbf{f}$ and $\mathbf{f}^+$, which makes intuitive sense and further reveals contrastive disentanglement relax similarity constrains to be in feature domain rather than in image domain.

\begin{figure}[t]
	\centering
	\subfloat[]{%
		\includegraphics[width=0.165\textwidth]{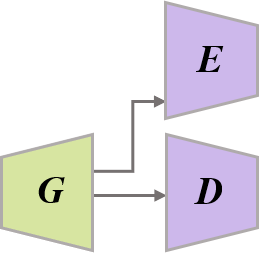}%
		\label{fig:train_config_a}%
	}\hspace{0.6cm}
	\subfloat[]{%
		\includegraphics[width=0.165\textwidth]{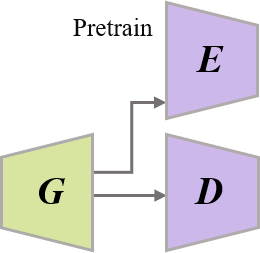}%
		\label{fig:train_config_b}%
	}\hspace{0.6cm}
	% \subfloat[]{%
	% 	\includegraphics[width=0.165\textwidth]{Figs/fewlabels.png}%
	% 	\label{fig:train_config_c}%
	% }\hspace{0.6cm}
	% \subfloat[]{%
	% 	\includegraphics[width=0.165\textwidth]{Figs/Pretrain_few.png}%
	% 	\label{fig:train_config_d}%
	% }
	\caption{\textbf{Comparison of several extensions of the original \texttt{CoDeGAN}.} (a): \texttt{CoDeGAN}. (b): \texttt{CoDeGAN} (self) or \texttt{CoDeGAN} (meta).  $E_c$ is pretrained by contrastive learning or meta-learning. 
 % (c): \texttt{CoDeGAN} + Few Labels. $E_c$ is learned under the guidance of few labeled real images. (d): \texttt{CoDeGAN} (self) or \texttt{CoDeGAN} (mata) + Few Labels. $G$, $D$, and $E_c$ are all pretrained, and $E_c$ is pretrained by contrastive learning or meta-learning. Meanwhile, $E_c$ is learned under the guidance of few labeled images.
 }
	\label{fig:train_config}
\end{figure}
%-------------------------------------------------------------------------
\section{Pre-Training Class Related Encoder}
\label{DisPrior}

In CL~\citep{tschannen2019mutual,arora2019theoretical}, the learned representation is always helpful for downstream tasks.
In~\citep{arora2019theoretical}, the authors demonstrated that the unsupervised loss function $L_{un}\left(f\right)$ for contrastive learning could be a surrogate for the downstream supervised tasks,
	\begin{equation}
		L_{sup}\left(f\right) \leq \alpha L_{un}\left(f\right), \forall f\in\mathcal{F},
	\end{equation}
where $f$ denotes the representation function, $L_{sup}\left(f\right)$ denotes the downstream supervised loss (\ie~average classification loss).
Here $\alpha \to 1$ when the number of the latent class is infinite, and the classes occur uniformly in the unlabeled data.
This formulation tells us that the representation learned by contrastive learning is semantic. 
Intuitively, the learned semantic representation can enforce the network to learn meaningful factors. 
It means contrastive pre-training (which can be unsupervised or supervised) can learn prior knowledge as guidance for disentanglement.

%\section{Introducing Few Labels}
%This subsection details the few-labels setting for \texttt{CoDeGAN} in Sec.5.
%Locatello~\etal~\citep{locatello2019challenging} recently demonstrated that unsupervised disentanglement learning without inductive biases is theoretically impossible.
%To incorporate limited supervision, our work uses few labeled images as anchor points.
%For this purpose, we modify the contrastive loss $\mathcal{L}_c$ in (\textcolor{red}{4}) of the main paper as bellows,
%\begin{equation}
%    e^{\left(\mathbf{f}^T\mathbf{f}^+/\tau\right)}
%    +e^{\left(\mathbf{f}^T\mathbf{f}_{real}^{+}/\tau\right)}.
%\end{equation}
%where $\mathbf{f}^+_{real}$ the representations of real images.
%Thus, for each generated image, we encourage its representations to be close to one anchor point, which may make use of these few labels to the largest extent. 
%Similarly, the contrastive loss for negative pairs can have some samples from real images with different labels.
% Fig.~\ref{fig:train_config_b} and Fig.~\ref{fig:train_config_d} show CoDeGANs which are trained with few labels.
%Fig.~\ref{fig:train_config_c} and Fig.~\ref{fig:train_config_d} show \texttt{CoDeGAN}s, which are trained with few labels.
%Thus, the contrast here does not only means contrast with positive and negative pairs but also means comparing fake examples with anchor points (\ie, real images with few labels), which may largely promote the disentangling performance, even if few labels are provided.

%-------------------------------------------------------------------------
\section{Experimental Results}

We conduct extensive experiments to assess the effectiveness of our proposed method on widely-used datasets. In the subsequent section, we present the disentanglement accuracy and generative quality of \texttt{CoDeGAN} alongside other baseline approaches.

\subsection{Experimental Setup}

\noindent\textbf{Datasets.}  The MNIST~\citep{lecun2001gradient}, Fashion-MNIST~\citep{lecun2001gradient}, CIFAR-10~\citep{Krizhevsky2009LearningML}, COIL-20~\citep{nenecolumbia}, 3D-Cars~\citep{fidler20123d}, and 3D-Chairs~\citep{aubry2014seeing} datasets have been widely and consistently used in disentanglement, and they have greatly facilitated progress in the community.
The details of these datasets are given below:
(\romannumeral1) MNIST~\citep{lecun2001gradient} has $70,000$ $28\times28$ grayscale images of digits ranging from 0 to 9.
(\romannumeral2) Fashion-MNIST~\citep{xiao2017/online} has $70,000$ images with the size $28\times28$.
(\romannumeral3) CIFAR-10~\citep{Krizhevsky2009LearningML} has $60,000$ labeled $32\times32$  RGB natural images in $10$ classless.
(\romannumeral4) 3D-Cars~\citep{fidler20123d}. It consists of $183$ categories, and each category has $96$ images. We followed~\citep{ojha2020elastic} to randomly sample $10$ categories ($960$ RGB images) images, and all images were resized to $64\times64$.
(\romannumeral5) 3D-Chairs~\citep{aubry2014seeing}. It consists of 1396 total categories, and each category has 62 images with varying positions. We followed~\citep{ojha2020elastic} to randomly sample $10$ categories ($620$ RGB images) images, and all images were resized to $64\times64$.
(\romannumeral6) COIL-20~\citep{nenecolumbia}. It has 1440 natural images in $20$ classes. Following~\citep{mukherjee2019clustergan}, all images were resized to $32\times32$.

%-------------------------------------------------------------------------
\noindent\textbf{Evaluation metrics.} 
The two criteria for disentanglement model evaluation are disentanglement accuracy and generative quality.
Thus, normalized clustering purity (ACC), normalized mutual information (NMI), and adjusted rank index (ARI) are used to evaluate disentanglement accuracy~\citep{mukherjee2019clustergan}.
Inception Score (IS)~\citep{salimans2016improved} and Fr\'{e}chet Inception Distance (FID)~\citep{heusel2017gans} are used to evaluate image quality.
% \jw{We don't choose MIG and DCI measure GAN's ability to learn a disentangled representation because is which rely on qualitative assessments, and the Factor VAE score of discrete GAN-based methods is almost close to 1.}
The details of these metrics are given below:
(\romannumeral1) ACC is designed to calculate the percentage of correct clusters in total samples.
(\romannumeral2) NMI tends to evaluate cluster score by measuring the normalized mutual information between real and clustered labels.
(\romannumeral3) ARI is a metric that measures the similarity of the two clusters by evaluating the proportion of the correct clustering in all clustering results.
Following the work~\citep{mukherjee2019clustergan}, given test images, $k$-means algorithm is adopted to cluster encoded representations, and the best ACC, NMI, and ARI are reported from 5 times run.
(\romannumeral4) IS is calculated by using a pre-trained Inception v3 model to predict the class probabilities for each generated image.
(\romannumeral5) FID compares the distributions of Inception embeddings of real data distribution and model distribution. 
In this work, IS is calculated for ten partitions of $50,000$ randomly generated images, and FID is calculated for $50,000$ images.

 Due to space limitations, more implementation details of our experiment are in \Cref{sec:idetails}.

\begin{figure}[ht]
\centering
% \begin{subfigure}{0.45\textwidth}
% \centering
% \includegraphics[width=\linewidth]{./Figs/fig3_v4.pdf}\subfloat[$L_4$]{\includegraphics[width=0.47\linewidth]{Cifar10_CoDeGAN_multi_L4_1.png}}\hspace{0.2cm}
\subfloat[Variations in ACC and FID with different $\beta_1$]{\includegraphics[width=\linewidth,page=1]{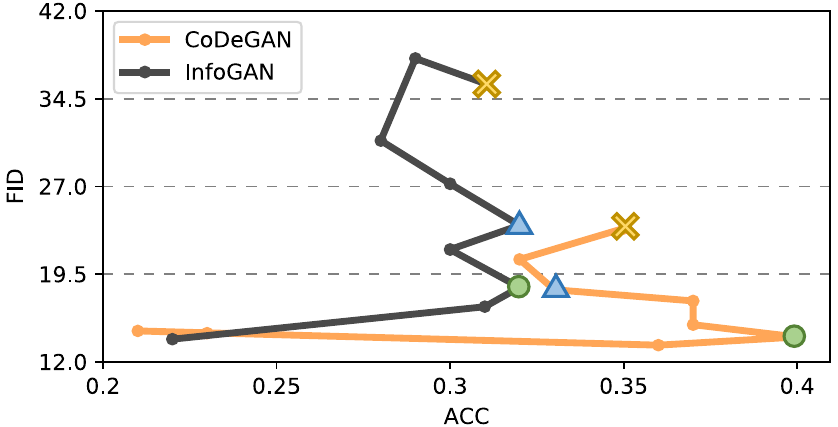}\label{fig:acc-fid}}\\
% \subfloat[ACC-FID 2D diagram]{\includegraphics[width=\linewidth,page=2]{Figs/fig3_v7.pdf}}\\
\subfloat[InfoGAN]{\includegraphics[width=\linewidth,page=2]{Figs/fig3_v6.pdf}\label{fig:infogan_acc}}\\
\subfloat[CoDeGAN]{\includegraphics[width=\linewidth,page=3]{Figs/fig3_v6.pdf}\label{fig:codegan_acc}}\\
% \includegraphics[width=0.45\textwidth]{./Figs/Fig_ACC_FID_InfoGAN_CoDeGAN_v2.pdf}
% \subcaption*{CoDeGAN}
% \label{fig:ACCFID-cifar10-a}
% \end{subfigure}
\caption{\textbf{Disentanglement accuracy and generative quality of \texttt{CoDeGAN} and InfoGAN with different trade-offs on CIFAR-10.} (a) X-axis denotes ACC ($\%$), and y-axis denotes FID. \textbf{Orange line:} \texttt{CoDeGAN} with loss $\mathcal{L}_{GAN} + \beta_1 \mathcal{L}_c$. \textbf{Black line:} InfoGAN with loss$\mathcal{L}_{GAN}+\beta_1\mathcal{L}_{MI}$.
CodeGAN achieves higher ACC and lower FID than InfoGAN for most the trad-offs.
(b) The worst generated images of CoDeGAN with different trade-offs.
(c) The worst generated images of InfoGAN with different trade-offs.
From left to right, the generated images corresponds to the best(green circle), the much lager(blue triangle), and the largest(yellow cross) $\beta_1$.
The presence of red boxes in (b) and (c) indicates mode collapse for some certain class.}
\label{fig:ACCFID}
\end{figure}
	
% \begin{figure*}[h]
% 	\centering
%         \subfloat{
%         \includegraphics[width=0.3\textwidth]{UnSupervised-Fashion-MNIST-tsne.png}
%         \label{fig:tsneunsup}
%         }
%         \subfloat{
%             \includegraphics[width=0.3\textwidth]{fashion_unsupervised_pretrained.png}
%             \label{fig:tsneself}
%         }
%         \subfloat{
%             \includegraphics[width=0.3\textwidth]{Supervised-Fashion-MNIST-tsne.png}
%             \label{fig:tsnefewlabels}
%         }
% 	\caption{\textbf{Visualization \jw{of the encoded features} of the generated images by variant \texttt{CoDeGANs}.} For all settings, $10000$ points are sampled from $p\left(\mathbf{z},c\right)$, the number of sampled points for each fixed $c$ is the same, 
%     \jw{and different color corresponds to different values of factor $c$.}
%     \textbf{(a):} \texttt{CoDeGAN}. \textbf{(b):} \texttt{CoDeGAN} with pre-trained $E_c$.  \textbf{(c):} \texttt{CoDeGAN} with pre-trained $E_c$ and little supervising.
% \jw{The encoder $E_c$ is pre-trained using SimCLR, and few labels means $0.2\sim 1\%$ train datasets have the labels.}}
% 	\label{fig:tsne}
% \end{figure*}	
\begin{figure}[h]
\centering
    \subfloat[]{
    \includegraphics[width=0.47\linewidth]{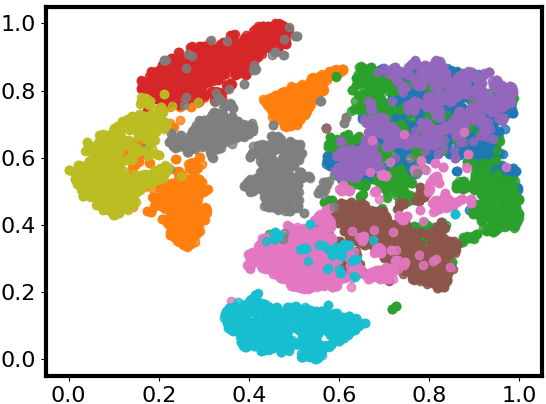}
    \label{fig:tsneunsup}
    }
    \subfloat[]{
        \includegraphics[width=0.47\linewidth]{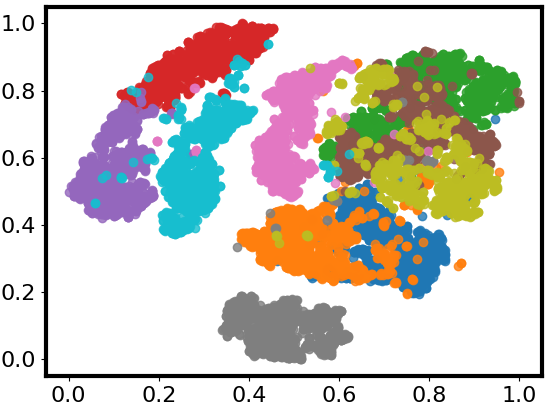}
        \label{fig:tsneself}
    }
\caption{\textbf{Visualization of the encoded features of the generated images by variant \texttt{CoDeGANs}.} For all settings, $10000$ points are sampled from $p\left(\mathbf{z},c\right)$, the number of sampled points for each fixed $c$ is the same, 
and different color corresponds to different values of factor $c$.
(a): \texttt{CoDeGAN}. (b): \texttt{CoDeGAN} with pre-trained $E_c$. 
The encoder $E_c$ is pre-trained using SimCLR.}
\label{fig:tsne}
\end{figure}	

%--------------------------------------------------------------------------
\subsection{Disentanglement Performance against Different Trade-offs}

To demonstrate \texttt{CoDeGAN} enhances generation performance and disentanglement accuracy , we conduct a comparative analysis between \texttt{CoDeGAN}, employing only the contrastive loss $\mathcal{L}_c\left(\mathbf{z},c\right)$, and InfoGAN. Our comparison encompasses disentanglement accuracy and generation quality. Specifically, we evaluate the ACC and FID values of \texttt{CoDeGAN} and InfoGAN across various trade-offs. \Cref{fig:acc-fid} illustrates the ACC and FID curves on CIFAR-10. \Cref{fig:infogan_acc} and \Cref{fig:codegan_acc} illustrate the worst image generated by InfoGAN and \texttt{CoDeGAN}.

It is apparent that increasing $\beta_1$ in \texttt{CoDeGAN} worsens FID slowly, while increasing $\beta_1$ in InfoGAN decreases FID dramatically.
This demonstrates that the contrastive loss, due to alleviating GAN's training instability, would not dramatically destroy generative quality even when increasing $\beta_1$, thus could leave more room to constrain similarity.
One could observe that \texttt{CoDeGAN} is located in the lower right corner of the figure, which means better performance.

To visualize the generation performance of InfoGAN and \texttt{CoDeGAN} with different $\beta_1$, we display their generated images in \Cref{fig:infogan_acc} and \Cref{fig:codegan_acc}.
Each row corresponds to the best, the larger, and the largest $\beta_1$ values.
In addition, the worst cases are chosen for display.
We observe \texttt{CoDeGAN} has better disentangling results and generation quality compared to InfoGAN.
When $\beta_1$ is slightly increased, InfoGAN experiences mode collapse while CoDeGAN maintains stable generation.
When $\beta_1$ is significantly increased, InfoGAN suffers from mode collapse across all classes and exhibits poor generatoin quality while CoDeGAN only experiences mild mode collapse.

%--------------------------------------------------------------------------
\subsection{Disentanglement Performance with Pre-Training }

\noindent\textbf{Effects of pre-training.} 	
To validate the effectiveness of self-supervised pre-training in disentanglement, we compare \texttt{CoDeGAN} with and without pre-trained encoder on Fashion MNIST.  
\Cref{fig:tsne} demonstrates that pre-training can learn valid priors as guidance for unsupervised disentanglement. 
In \Cref{fig:tsneunsup}, the `green', `pink', and `purple' points in the t-SNE plot are tangled, while the `gray' and `orange' points are not in the same clusters.
% Fig.~\ref{fig:tsneself} and Fig.~\ref{fig:tsnefewlabels} show contrastive pre-training and few labels aids to promote disentanglement performance, where the points with the same color become more close.
\Cref{fig:tsneself} shows that contrastive pre-training aids to promote disentanglement performance, where points with the same color become more close.

\begin{figure}
\centering
    \subfloat[Fashion-MNIST]{
    \includegraphics[ height=3cm]{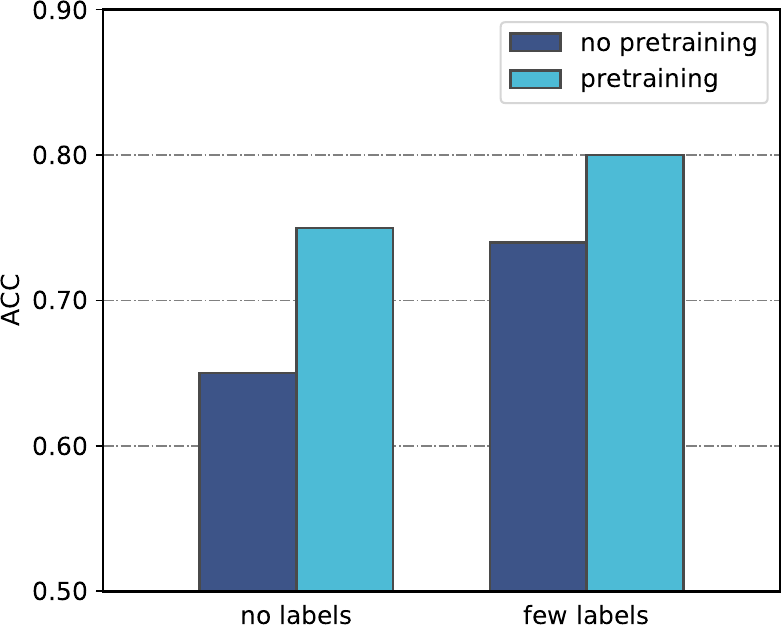}
    \label{fig:pretraining_Fashion}
    }
    \subfloat[3D-Chairs, 3D-Cars and COIL-20]{
    \includegraphics[height=3cm]{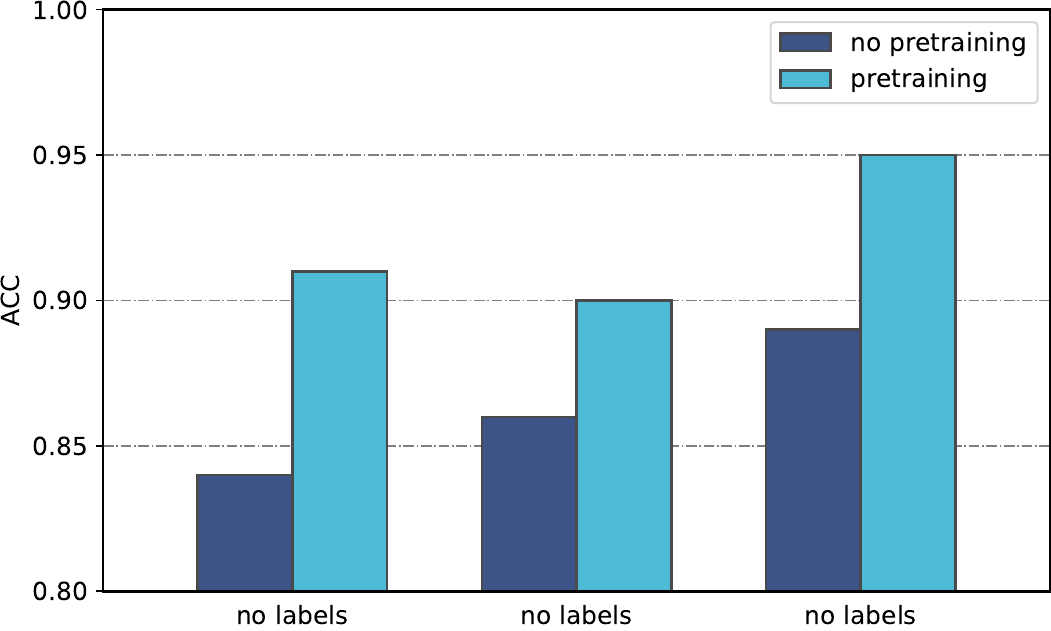}
    \label{fig:pretraining_CIFAR}
    }
 	\caption{\textbf{Disentanglement accuracy of \texttt{CoDeGAN} with and without pre-training.} Pre-training significantly improves the disentanglement accuracy on Fashion-MNIST, 3D-Chairs, 3D-Cars and COIL-20. }
 	\label{fig:box-plot}
 \end{figure}

The results in \Cref{fig:box-plot} further support the idea of integrating pre-training into disentanglement.
As can be seen from \Cref{fig:pretraining_Fashion,fig:pretraining_CIFAR}, utilizing the pre-training encoder gains improvement by $10\%$ in ACC ($0.65\rightarrow0.75$) on Fashion-MNIST, $9\%$ in ACC $\left(0.84\rightarrow0.93\right)$ on 3D-Chairs, and $8\%$ in ACC $\left(0.86\rightarrow0.94\right)$ on 3D-Cars.
These are somewhat surprising findings.

\begin{figure}[ht]
 	\centering
            \subfloat{\includegraphics[width=0.23\textwidth]{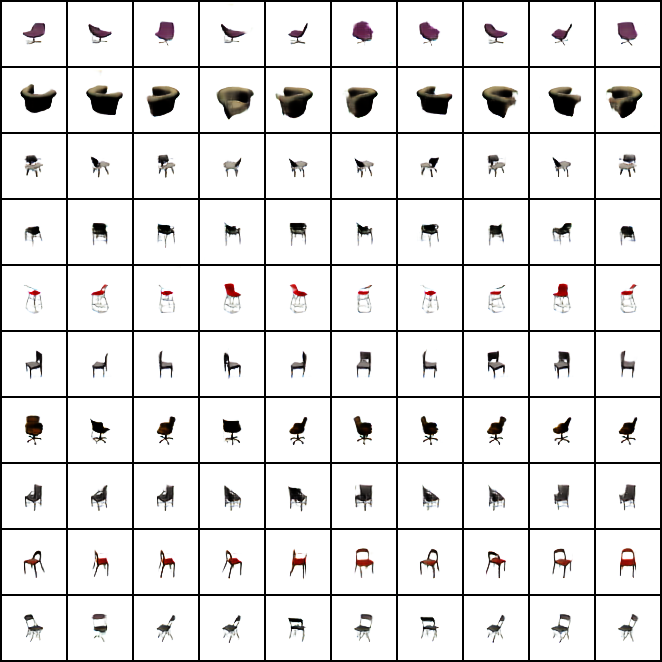}}
		  \vspace{1.5pt}
            \subfloat{\includegraphics[width=0.23\textwidth]{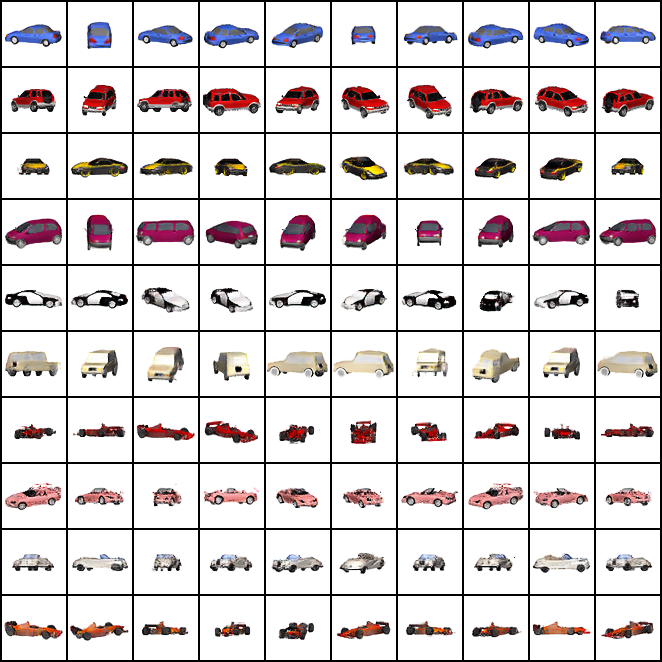}}
	\caption{\textbf{Example images generated by \texttt{CoDeGAN} for 3D-Chairs and 3D-Cars.} Each row corresponds to one category object with varying pose. The encoder $E_c$ is pre-trained using self-supervised pre-training method\textemdash SimCLR.}
	\label{fig:chairs_cars}
\end{figure}

\begin{figure*} % Requires 
	\centering         
            \subfloat{\includegraphics[width=0.3\textwidth]{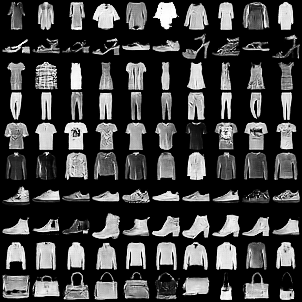}}
            \vspace{2pt}
            \subfloat{\includegraphics[width=0.3\textwidth]{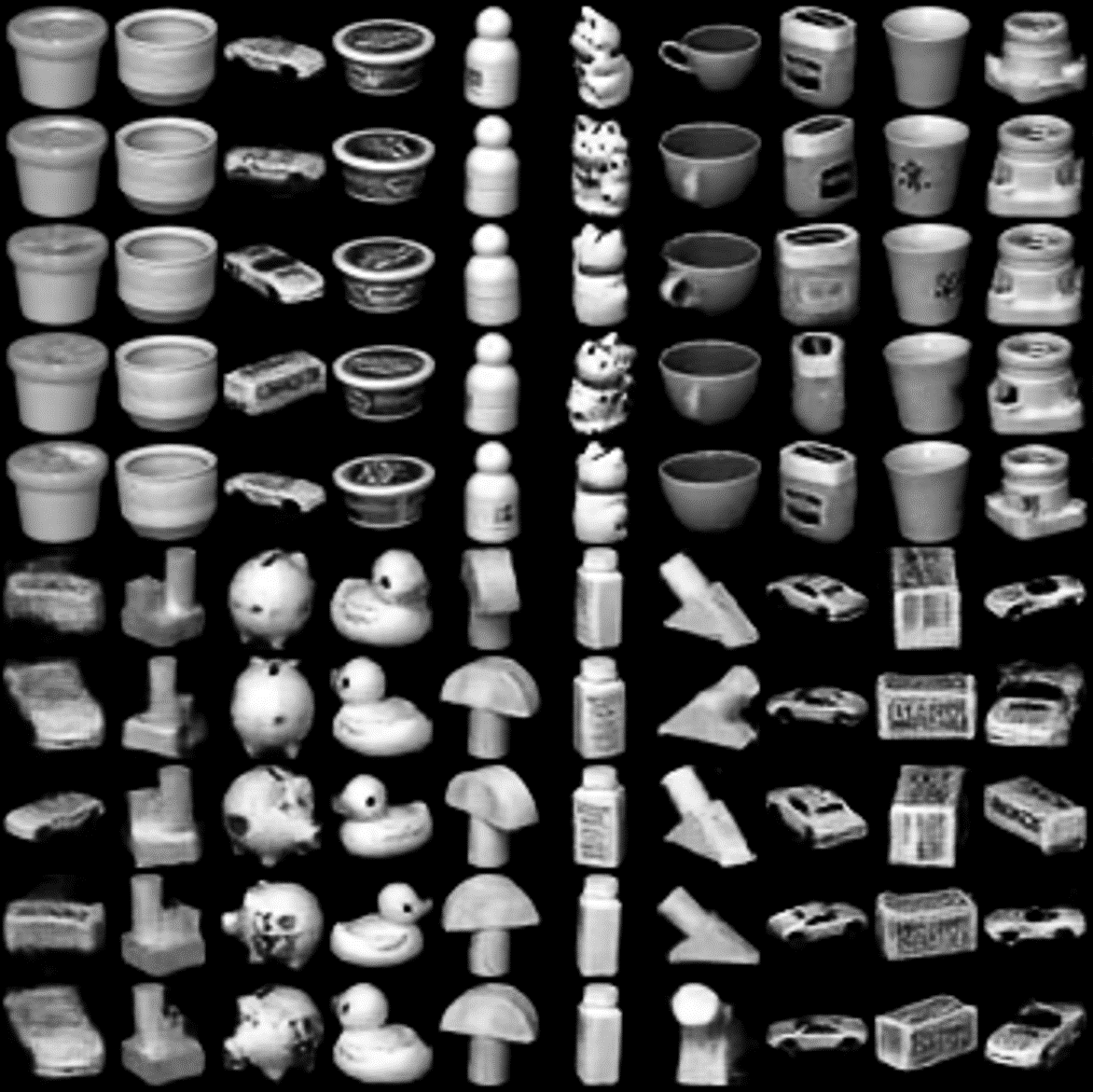}}
            \vspace{2pt}
            \subfloat{\includegraphics[width=0.3\textwidth]{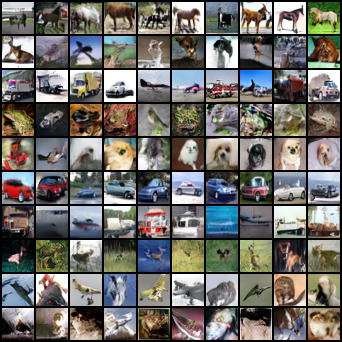}}
	\caption{\textbf{Visualization of the images generated by  $\texttt{CoDeGAN}$ (pre-trained) for Fashion-MNIST, COIL-20 and CIFAR-10.}}
	\label{fig:result_fig}
\end{figure*}

In addition, some representative images generated by \texttt{CoDeGAN}, with pre-trained encoder $E_c$, on 3D-Chairs and 3D-Cars are shown in \Cref{fig:chairs_cars}.
Some more results on Fashion-MNIST, COIL-20, and CIFAR-10 are shown in \Cref{fig:result_fig}.
These results qualitatively confirm that the pre-trained \texttt{CoDeGAN} has the potential to disentangle image variation unsupervisedly.

% \vspace{-0.2cm}

\noindent\textbf{Pre-training methods study.} 
A potential concern for pre-training is that different pre-training methods could affect the disentanglement accuracy of $\texttt{CoDeGAN}$.
To evaluate this, we compare three recently proposed pre-training methods on the 3D-Chairs, COIL-20, and Fashion-MNIST datasets, including SimCLR~\citep{hinton2020SIMCLR}, SimSiam~\citep{chen2021Siamese} and NNCLR~\citep{dwibedi2021little}.

\Cref{pretrain_methods} reports the comparison results.
For example, SimCLR's ACC, NMI and ARI values are $0.91, 0.90, 0.85$ on 3D-Chairs, $0.95, 0.95, 0.92$ on COIL-20, and $0.75, 0.72, 0.63$ on Fashion-MNIST.
The reported results reveal the disentanglement accuracy, \ie, ACC, of SimCLR, SimSiam and NNCLR only has slight difference $(\text{in}~3\%, ~5\%,~\text{and}~3\%)$ on 3D-Chairs, COIL-20 and Fashion-MNIST.
These results suggest that different pre-training methods do not distinctly affect the disentanglement accuracy.

% \begin{table}[h]
% 	\centering
% 	\resizebox{0.5\textwidth}{!}{
% 		\begin{tabular}{|c|ccc|ccc|ccc|}
% 			\hline
% 			&\multicolumn{3}{c|}{SimCLR~\citep{finn2017model}}
% 			&\multicolumn{3}{c|}{SimSiam ~\citep{hinton2020SIMCLR}}
% 			&\multicolumn{3}{c|}{ NNCLR~\citep{dwibedi2021little}}\\
% 			%\cline{2-10}
% 			&ACC$\uparrow$&NMI$\uparrow$&ARI$\uparrow$ &ACC$\uparrow$&NMI$\uparrow$&ARI$\uparrow$
% 			&ACC$\uparrow$&NMI$\uparrow$&ARI$\uparrow$\\
% 			\hline
% 			\hline
% 			3D-Chairs    &0.91   & 0.90 & 0.85 & 0.93 & 0.92  & 0.89 &0.90 &0.92 &0.87\\
% 			COIL-20      &0.95   & 0.95 & 0.92 & 0.90  & 0.93 & 0.86 &0.93  &0.95&0.90\\
% 			FashionMNIST&0.75   & 0.72 & 0.63 & 0.72 & 0.67  & 0.58 &0.74 &0.69 &0.61\\
% 			\hline
% 		\end{tabular}
% 	}
% 	\caption{\textbf{Pre-training Methods Study}. We pre-train encoder $E_c$ with different  pre-trainng methods, which could be supervised or unsupervised.}
% \end{table}
\begin{table*}[ht]
    \caption{\textbf{Quantitative comparison of different pre-training methods.} Different pre-training methods do not affect the disentanglement accuracy distinctly.}
    \centering
% 	\resizebox{0.45\textwidth}{!}{
    % \scriptsize
    % \setlength{\tabcolsep}{2pt}
		\begin{tabular}{lccccccccc}
            \toprule
            % \hline
			\multirow{2}[1]{*}{\textbf{Datasets}}&\multicolumn{3}{c}{SimCLR~\citep{finn2017model}}
			&\multicolumn{3}{c}{SimSiam ~\citep{hinton2020SIMCLR}}
			&\multicolumn{3}{c}{NNCLR~\citep{dwibedi2021little}}\\
			\cmidrule(lr){2-4}\cmidrule(lr){5-7}\cmidrule(lr){8-10}
			%\cline{2-10}
			&ACC$\uparrow$&NMI$\uparrow$&ARI$\uparrow$ &ACC$\uparrow$&NMI$\uparrow$&ARI$\uparrow$
			&ACC$\uparrow$&NMI$\uparrow$&ARI$\uparrow$\\
            \midrule
            % \hline
            % \hline
            % \midrule
			3D-Chairs    &0.91   & 0.90 & 0.85 & 0.93 & 0.92  & 0.89 &0.90 &0.92 &0.87\\
			COIL-20      &0.95   & 0.95 & 0.92 & 0.90  & 0.93 & 0.86 &0.93  &0.95&0.90\\
			Fashion-MNIST&0.75   & 0.72 & 0.63 & 0.72 & 0.67  & 0.58 &0.74 &0.69 &0.61\\
            \bottomrule
            % \hline
		\end{tabular}

	\label{pretrain_methods}
\end{table*}

\subsection{Comparison with Baselines}

\begin{table*}[ht]
\centering
		\caption{\textbf{Qualitative comparison with state-of-the-art methods on the Fashion-MNIST, COIL-20 and CIFAR-10 datasets.} $*$: Trained using open source code. }
		% \resizebox{0.45\textwidth}{!}{	
			\begin{tabular}{lccccc}%%%The number of columns has to be defined here
				% \hline
    \toprule
				Algorithm & ACC$\uparrow$ & NMI$\uparrow$ & ARI$\uparrow$& IS$\uparrow$ &FID$\downarrow$\\ %%%% Table body
				% \hline
    \midrule
				% \hline
				\multicolumn{6}{c}{\textbf{\textit{Fashion-MNIST}}}\\
				% \hline
    \midrule
				% \hline
				ClusterGAN~\citep{mukherjee2019clustergan} & $0.63$  & $0.64$ & $0.50$ &-&- \\
				InfoGAN~\citep{chen2016infogan}            & $0.61$  & $0.59$ & $0.44$ &-&-\\
				$\text{InfoGAN-CR}^*$~\citep{lin2020infogan}& $0.64$ & $0.64$ &$0.50$  &4.20&30.35\\
				GAN with bp ~\citep{mukherjee2019clustergan}&$0.56$  & $0.53$ & $0.37$ &-&-\\
				GAN with Disc.$\phi$~\citep{mukherjee2019clustergan} &$0.43$  & $0.37$ & $0.23$ &-&-\\
				\rowcolor{mypink}
				\texttt{CoDeGAN}       & $0.65$    &$0.66$  &  $0.52$   &$\textbf{4.36}$&$\textbf{17.09}$\\
				\rowcolor{mypink}
				\texttt{CoDeGAN} (self)& $\textbf{0.75}$ &$\textbf{0.72}$&  $\textbf{0.63}$   &$4.18$&$19.51$\\
				% \hline
    \midrule
				\multicolumn{6}{c}{\textbf{\textit{COIL-20}}}\\
				% \hline
    \midrule
				 $\text{ClusterGAN}^*$~\citep{mukherjee2019clustergan} & $0.82$ & $0.89$ & $0.79$ &$4.56$&$48.13$\\
				$\text{InfoGAN}^*$~\citep{chen2016infogan}    & $0.85$  & $0.90$ & $0.81$ &$4.58$&$64.49$\\
				$\text{InfoGAN-CR}^*$~\citep{lin2020infogan} & $0.85$ & $0.90$ & $0.82$ &$\textbf{4.96}$&$69.71$\\
				\rowcolor{mypink}
				\texttt{CoDeGAN}    & $0.89$    &$0.91$  &  $0.84$   &$4.54$&$\textbf{45.63}$\\
				\rowcolor{mypink}
				\texttt{CoDeGAN} (self)    & $\textbf{0.95}$    &$\textbf{0.95}$  &  $\textbf{0.92}$   &$4.61$&$46.18$\\
				% \hline
    \midrule
				% \hline
				\multicolumn{6}{c}{\textbf{\textit{CIFAR-10}}}\\
				% \hline
    \midrule
				% \hline
				$\text{ClusterGAN}^*$~\citep{mukherjee2019clustergan} &$0.18$ &0.05 &0.03 &$8.18$ & $14.95$\\
				$\text{InfoGAN}^*$~\citep{chen2016infogan} & $0.31$ &0.21  &0.12  & $7.89$  & $16.71$\\
				% $\text{InfoGAN-CR}^*$~\citep{lin2020infogan}& 0.35  &0.25  &0.18  & $7.35$  & $21.58$\\
				$\text{InfoGAN-CR}^*$~\citep{lin2020infogan}& 0.34  &0.24  &0.15  & $7.70$  & $25.19$\\
				\rowcolor{mypink}
				\texttt{CoDeGAN} & $0.44$ &$0.35$  &$0.26$&$8.29$& $14.17$\\
				\rowcolor{mypink}
				\texttt{CoDeGAN} (self) & \textbf{0.50} &\textbf{0.35}&\textbf{0.27} &\textbf{8.33} &\textbf{13.70}\\
				% \hline
    \bottomrule
			\end{tabular}
		% }
		\label{tab:no_label}
%\vspace{0.2cm}
\end{table*}
\noindent\textbf{Disentanglement accuracy comparison.} 	To compare with prior work, we conduct testing on Fashion-MNIST, COIL-20, and CIFAR-10.
Here, `self’ indicates the encoder in \texttt{CoDeGAN} is pre-trained by SimCLR~\citep{hinton2020SIMCLR}.
%, and `meta' indicates the encoder is pre-trained by MAML~\citep{finn2017model}.
The comparison results in \Cref{tab:no_label} showcase consistent tendencies.
(\romannumeral1) Unsupervised \texttt{CoDeGAN} achieves SOTA performance on multiple benchmarks.
On the challenging CIFAR-10 dataset, \texttt{CoDeGAN} gains a stunning 19\% absolute improvement over InfoGAN and a 16\% absolute improvement over the previous SOTA methods.
(\romannumeral2) Contrastive pre-training can learn prior knowledge to guide disentanglement in unsupervised settings.
It gains improvement by $10\%$ in ACC on Fashion-MNIST, and $6\%$ on CIFAR-10, compared to \texttt{CoDeGAN} without pre-training. 
% It further gains improvement by $6\%$ in ACC on FashionMNIST, and $8\%$ on CIFAR-10 (with few labels). 
Considering that SimCLR evaluated using KNN has an accuracy of $0.55$ on CIFAR-10 and $0.69$ on Fashion-MNIST, it is not surprising that it provides a huge boost to unsupervised disentagnlement.
%(\romannumeral3) Few labels (usually $0.2\sim 1\%$) can dramatically promote the disentanglement performance. 

Disentangling category variations of natural images is extremely challenging as these images contain complex backgrounds, object shapes, and appearances.
Most unsupervised methods fail on this task, such as disentangling on CIFAR-10.
As seen from \Cref{tab:no_label}, ClusterGAN only has an ACC of 0.18 on CIFAR-10.
It is noteworthy that this result is much lower than InfoGAN. 
A possible explanation for this might be that the encoder of ClusterGAN is separated from the discriminator, leading to learned representations that may not be semantically related to the image category. 
In contrast, after pre-training, \texttt{CoDeGAN} achieves an ACC of 0.50. 
The explanations are twofold: (\romannumeral1) The contrastive loss can improve GAN's equilibrium, leaving more room to constrain disentanglement by increasing the trade-off $\beta_1$. (\romannumeral2) The pre-trained encoder $E_c$ can guide \texttt{CoDeGAN} to learn semantic representations, enforcing the network to learn meaningful factors.

\noindent\textbf{Generative quality comparison.} \Cref{tab:no_label} also reports the IS and FID comparison results between our method and other baseline methods on different datasets.
For CIFAR-10, our method has the best IS and FID with $8.33$ and $13.70$ in unsupervised settings, respectively.
The results are also significantly better than InfoGAN and InfoGAN-CR.
As discussed previously, maximizing MI may limit diversity and break GAN's equilibrium.
In addition, contrastive regularization (CR) may further aggregate mode collapse, as shown in \Cref{tab:no_label}, where InfoGAN-CR has a FID of $25.19$ on CIFAR-10 compared to our result of $14.17$. 
\subsection{Ablation Study}

In our ablation study, we analyze the effect of constraining similarity on different feature layers, the effect of the contrastive loss, and the effect of reconstruction loss in \texttt{CoDeGAN}.

\noindent\textbf{Similarity constraint on different feature layer.}
As mentioned in \Cref{sec:ConDisA}, contrastive disentanglement relaxes the similarity constraints in the feature domain and is prone to improve GAN's training stability.
To further confirm this, we study the effects of contrastive loss on different feature layers.
The features from four layers of the encoder network $E_c$ at different resolutions $\left(L_1=32^2, L_2=16^2, L_3=8^2, L_4=4^2 \right)$ were extracted. 
Here, $L_4$ denotes the highest feature layer, and $L_1$ denotes the lowest feature layer, which is the most \emph{close} to the generated image.
The features were down-sampled four times as well as flattened. 
% Then, the multi-scale features were fed to a  ...
% Then, the multi-scale features $\left(L_1=4096, L_2=2048, L_3=1024, L_4=512 \right)$ were fed to a two-layer MLP, and reduced to the same dimensionality $(128)$.  
Then, the multi-scale features were fed to a two-layer MLP, and reduced to the same $128$ dimension.
%( 4096, 2048, 1024, 512 to 128 ).  
% reduced to the same dimensionality by MLP. 
%Following the work \cite{sauer2021projected}, we sum the contrastive loss on all feature layers. 

\Cref{multi_level} shows that $\texttt{CoDeGAN}$ has gradually worse IS and FID when adding contrastive loss $\mathcal{L}_c$ on feature layers that are close to the image.
We find the average FID scores for CIFAR-10 and 3D-Chairs fall to $24.27$ $\left(15.32\rightarrow24.27\right)$ and $79.10$ $\left(47.08\rightarrow79.10\right)$, and mode collapse sometimes occurs when adding contrastive loss on lower feature layers.
Specifically, after imposing contrastive loss from $L_1$ to $L_4$, the mode collapse rate changes as $0/5 \rightarrow 0/5 \rightarrow 1/5 \rightarrow 4/5$ for CIFAR-10. For 3D-Chairs, the collapse rate has advanced: $0/5 \rightarrow 0/5 \rightarrow 1/5 \rightarrow 2/5$. These findings suggest that imposing contrastive loss on lower feature layers might adversely affect model stability.

\begin{figure}[ht]
	\centering
	\subfloat[$L_4$]{\includegraphics[width=0.47\linewidth]{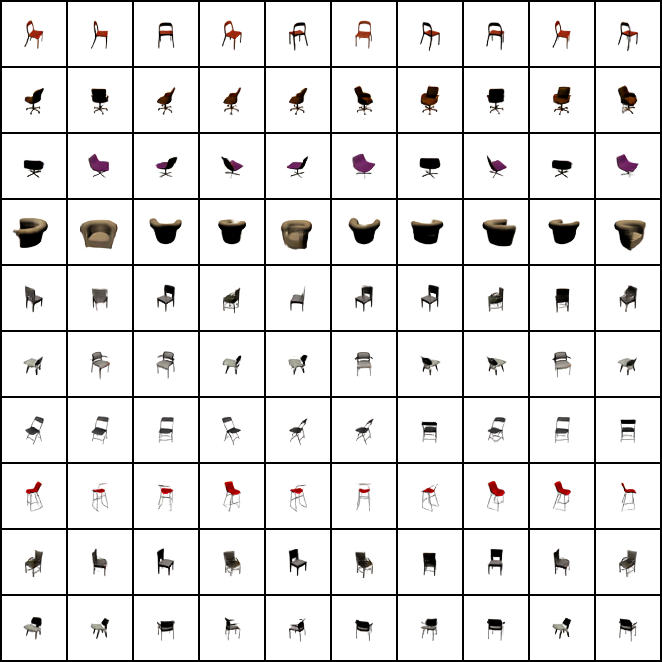}}\hspace{0.2cm}
	\subfloat[$L_3+L_4$]{\includegraphics[width=0.47\linewidth]{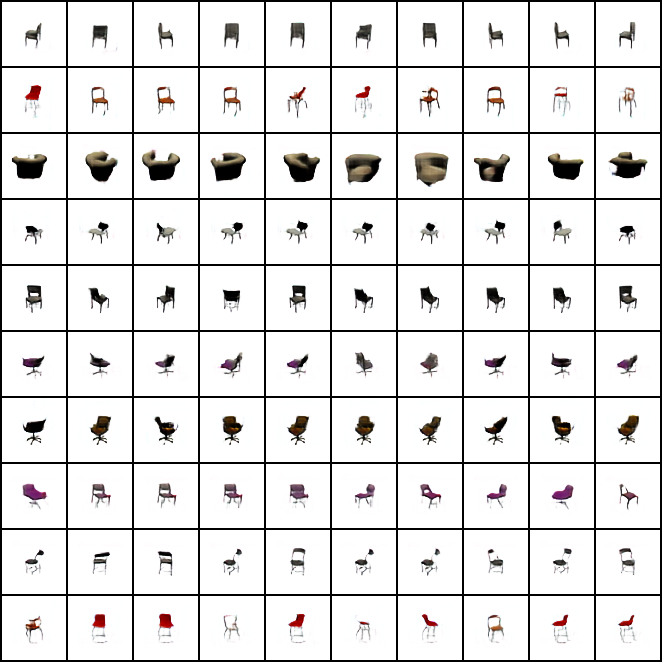}}\hspace{0.2cm}\\
	\subfloat[$L_2+L_3+L_4$]{\includegraphics[width=0.47\linewidth]{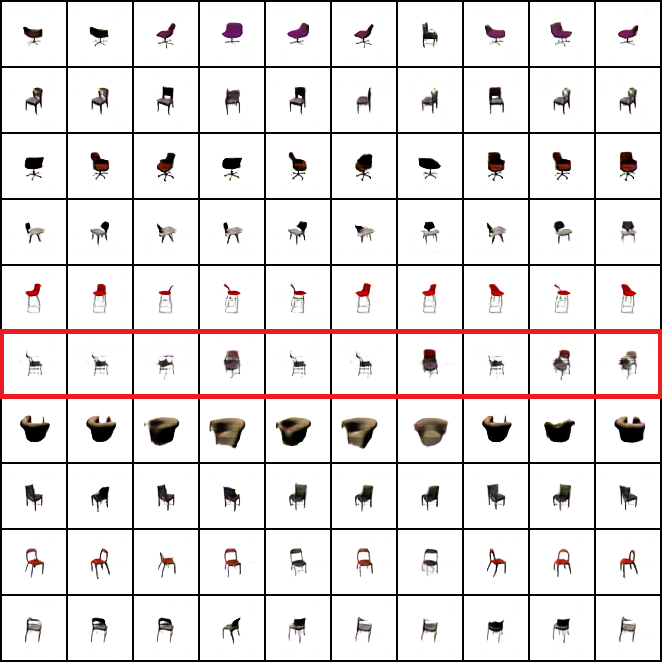}}\hspace{0.2cm}
	\subfloat[$L_1+L_2+L_3+L_4$]{\includegraphics[width=0.47\linewidth]{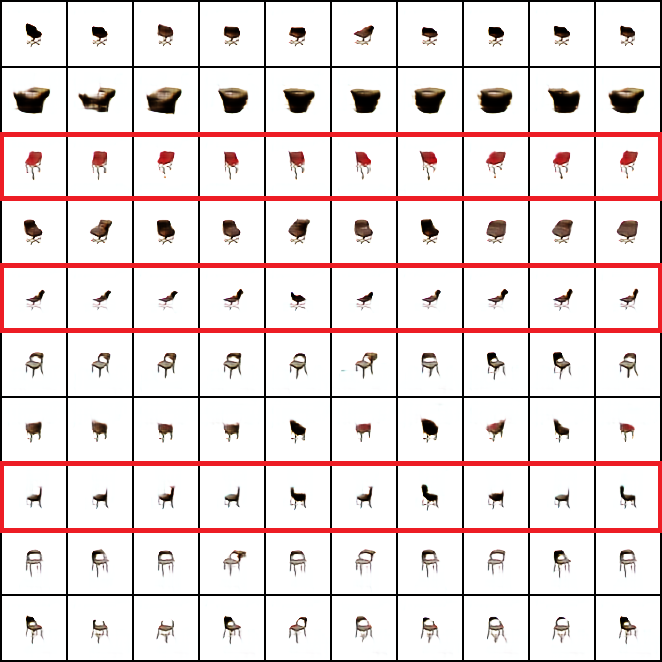}}
	\caption{\textbf{Visualization of generated images for 3D-Chairs with $L_c$ imposing on the different feature layers.} $L_4$ denotes the highest feature layer, and $L_1$ denotes the lowest feature layer, which is most \emph{close} to the input image. 
Each row is generated with same $c$ and different $\mathbf{z}$.
To facilitate model collapse analysis, we show the worst images from five training times}
	\label{fig:chair_multi}
\end{figure}

\Cref{fig:chair_multi} exhibits some generation results on the 3D-Chair dataset when applying $\mathcal{L}_c$ on different layers.
Similar to the observations on CIFAR-10, imposing $\mathcal{L}_c$ on lower layers leads to more obvious mode collapse happens.
For generated images where $\mathcal{L}_c$ is applied to $\mathcal{L}_2+\mathcal{L}_3+\mathcal{L}_4$, it is observed that the generated images within red boxes, despite having different $\mathbf{z}$, appear to be identical.
When $\mathcal{L}_c$ is impoesed to the lowest layer, the frequency of mode collapse escalates.
More generated images with mode collapse can be found in \Cref{sec:appendix_abl}.
This tendency may be explained by the fact that the similarity constraints relaxing on low-layer features are slight.
This is because these layers are close to the image, and the model capacity of the low-layer sub-networks of the encoder $E_c$ is restricted. 

\begin{table}[h]
	\centering
 		\caption{\textbf{Disentanglement performance with different features in $\mathcal{L}_c$}. The results suggest that $\texttt{CoDeGAN}$ has gradually worse IS and FID when adding contrastive loss $\mathcal{L}_c$ on feature layers which are close to image.}
% 	  \scriptsize
    % \setlength{\tabcolsep}{2pt}
    		\resizebox{0.48\textwidth}{!}{	
		\begin{tabular}{ccccccccccc}
			
% 			\hline
            \toprule
			\multicolumn{4}{c}{\textbf{Layer}} & \multicolumn{1}{c}{\textbf{ACC$\uparrow$}} & \multicolumn{3}{c}{\textbf{IS$\uparrow$}} & \multicolumn{3}{c}{\textbf{FID$\downarrow$}} \\ 
			\cmidrule(lr){1-4} \cmidrule(lr){5-5}\cmidrule(lr){6-8}\cmidrule(lr){9-11}
			$L_4$ & $L_3$ & $L_2$ & $L_1$ & \multicolumn{1}{c}{Mean}  & \multicolumn{1}{c}{Max} & \multicolumn{1}{c}{Mean} & \multicolumn{1}{c}{Min} & \multicolumn{1}{c}{Max} & \multicolumn{1}{c}{Mean} & \multicolumn{1}{c}{Min} \\
% 			\hline
            \midrule
            % \midrule
			\multicolumn{11}{c}{\textbf{\textit{CIFAR-10}}}\\
            \midrule
			\checkmark&&& & 0.39   & 8.30 &  8.22 & 8.10 & 16.5 & 15.32 & 14.11 \\
			\checkmark&\checkmark&& & 0.40     & 8.42 & 8.11 & 7.90 & 16.97 & 15.86 & 14.78 \\
			\checkmark&\checkmark&\checkmark& & 0.42   & 8.32 & 7.98 & 7.47 & 26.62 & 17.31 & 12.74 \\
			\checkmark&\checkmark&\checkmark&\checkmark  & 0.43    & 8.15 & 7.67  & 6.86 & 37.41 & 24.27 & 13.92 \\
            \midrule
			\multicolumn{11}{c}{\textbf{\textit{3D-Chairs}}}\\
% 			\hline	
	
            \midrule
			\checkmark&&& & 0.84   & 3.60 &  3.45 & 3.08 & 77.13 & 47.08 & 35.98 \\
			\checkmark&\checkmark&& & 0.82     & 4.03 & 3.72 & 3.45 & 134.40 & 70.06 & 37.20 \\
			\checkmark&\checkmark&\checkmark& & 0.79   & 4.21 & 3.79 & 3.57 & 99.85 & 72.30 & 44.39 \\
			\checkmark&\checkmark&\checkmark&\checkmark  & 0.72    & 3.73 & 3.40  & 2.92 & 119.76 & 79.10 & 36.98 \\
			
% 			\hline	
			\bottomrule
		\end{tabular}}

		\label{multi_level}
	\end{table}

\begin{figure} % Requires 
	\centering 
	\begin{tabular}[h]{cc}
	    % \hspace{-0.4cm}
                \subfloat[]{\includegraphics[width=0.23\textwidth]{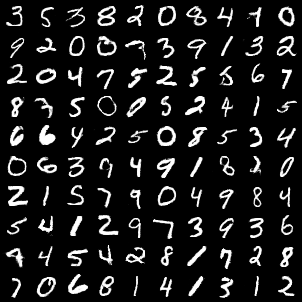}
                \label{fig:ablation_a}}
		&	
		\hspace{-0.7cm}	
                \subfloat[]{\includegraphics[width=0.23\textwidth]{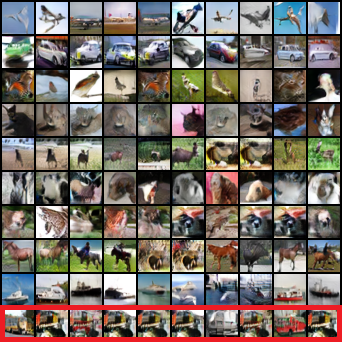}
                \label{fig:ablation_b}}
		\\
	    % \hspace{-0.4cm}

		% \begin{subfigure}[]{.25\textwidth}
        \subfloat[]{
			\centering
			\small
			% \resizebox{!}{0.65cm}{
			\begin{tabular}{c|ccc}
				% \hline
    \toprule
				$\mathcal{L}_c$ &  ACC $\uparrow$  & NMI$\uparrow$  &ARI$\uparrow$ \\
				% \hline
    \midrule
				% \hline
				W             &0.98& 0.96      & 0.96\\
				o              &0.11 & 0.00      & 0.00\\
				% \hline
    \bottomrule
			\end{tabular}
   
			\label{fig:ablation_c}
		}
		&
		\hspace{-0.7cm}
        \subfloat[]{
    		\centering
                \small
    		\begin{tabular}{c|ccc}
    			% \hline
       \toprule
    			$\mathcal{L}_\mathbf{z}$ &  IS$\uparrow$    & FID$\downarrow$& ACC$\uparrow$  \\
    			% \hline
    			% \hline
       \midrule
    			W              & 8.08   & 13.45&0.72\\
    			o               & 7.56   & 21.97&0.69\\
    			% \hline
       \bottomrule
    		\end{tabular}	
    		\label{fig:ablation_d}}\\
	\end{tabular}
	\caption{\textbf{Ablation study of contrastive loss $\mathcal{L}_c$ and reconstruction losses $\mathcal{L}_{\mathbf{z}}$.} (a) disentanglement result without $\mathcal{L}_c$ (b) generation result without $\mathcal{L}_{\mathbf{z}}$, (c) ACC, NMI and ARI scores with and without $\mathcal{L}_c$, (d) IS and FID scores with and without $\mathcal{L}_{\mathbf{z}}$.}
	\label{fig:ablation}
    \vspace{-5pt}
\end{figure}

\noindent\textbf{Effect of Contrastive loss.} To test the validity of contrastive loss $\mathcal{L}_c$ in terms of disentangling inter-class variation, we set the value of $\beta_1$ to be zero.
Then, the objective function only has two parts: (\romannumeral1) the loss $\mathcal{L}_{GAN}$, and (\romannumeral2) the loss $\mathcal{L}_\mathbf{z}$.
From the results in \Cref{fig:ablation_a} and \Cref{fig:ablation_c}, one can observe that there is no disentangling effect on MNIST, even when structured code including content factor $\mathbf{z}$ and class factor $c$ is fed into the network.
The ACC score falls to be 0.11 when $\beta_1=0$, whereas it rises to 0.98 when $\beta_1=75$.
Such results soundly demonstrate $\mathcal{L}_c$ can disentangle image category variation. 
	
\noindent\textbf{Effect of reconstruction loss.} 
To study the effect of reconstruction loss $\mathcal{L}_{\mathbf{z}}$, $\beta_2$ is set to be zero.
The worst results in 5 time runs are reported.
% On the CIFAR-10 dataset, the few labels setting with encoder pretrained by MAML~\citep{finn2017model} was randomly chosen for evaluation, where $\beta_1=0.25$.
\Cref{fig:ablation_b} and \Cref{fig:ablation_d} show the mode collapse image at 300 epoch and its corresponding IS and FID scores.
Mode collapse occurred two out of five times during the training process.
One can observe that the IS and FID scores fall to $7.56$ and $21.97$ when the mode collapse happens, while the IS and FID scores are $8.08$ and $13.45$ when no mode collapse happens.
This confirms taht the reconstruction loss can further alleviate the mode collapse problem.

% We study the features of different layers for contrastive disentanglement.
% The results are shown in Tab.~\ref{multi_level}.

%-------------------------------------------------------------------------
%\section{Discussion }
%Contrasitve disentanglement has some limitations.
%First, the original \texttt{CoDeGAN} suffers from the large variance on disentanglement scores, which are knows as common drawbacks of unsupervised disentanglement methods.
%Second, in unsupervised settings, \texttt{CoDeGAN} with contrastive pre-training is not guaranteed to perform better than original \texttt{CoDeGAN}, as the learned representation by unsupervised contrastive learning may be not semantic related.
%Third, \texttt{CoDeGAN} adopts contrative loss for disentanglement to alleviate mode dropping/collapse, which is a fundamental and long-standing problem in GANs. That does not mean \texttt{CoDeGAN} can avoid it completely.

%--------------------------------------------------------------------------
\section{Conclusion and Future Work}
We propose contrastive disentanglement for generative adversarial networks (\texttt{CoDeGAN}) in this paper, with both theoretical analysis and practical algorithmic implementation.
After relaxing the similarity constraint to the feature domain, \texttt{CoDeGAN} improves disentanglement accuracy and generation quality.
Integrating self-supervised pre-training to learn priors as guidance for disentanglement, \texttt{CoDeGAN} yields further improved disentanglement accuracy. 
We show that our proposed method achieves the SoTA performance on multiple benchmarks. 
In this paper, we only consider discrete factor disentanglement since it is harder to consider alleviating GAN's training instability and pre-training in continuous factor disentanglement.
Introducing pre-training techniques to multi-factor (which can be discrete or continuous) disentanglement is still an open problem.

\section*{Acknowledgment}
This work is partially supported by the National Natural Science Foundation of China (No.~62171111) and Natural Science Foundation of Sichuan Province (2023NSFSC1972).


\appendix

\section{Proofs}
\begin{lemma}
Minimizing the contrastive loss $\mathcal{L}_c$ is equivalent to maximizing the mutual information between generative representations $\mathbf{f}$ and $\mathbf{f}^+$, \ie, $E_c\left(G\left(\mathbf{z},c\right)\right)$ and $E_c\left(G\left(\mathbf{z}^+,c^+\right)\right)$.
% \begin{equation}
% \small
% \begin{split}
% 	\displaystyle \mathop{\mathbb{E}}
% 	_{\substack{\left(\mathbf{z}, c, \mathbf{z}^+, c^+\right)\sim p_{\text{pos}}\\ 
% 			\left(\mathbf{z}_i,c^-_i\right)\stackrel{i.i.d.}{\sim} p\left(\mathbf{z}, \text{neg}\left(c\right)\right)}}
% 	&\left[-\log \frac{ e^{\left(\mathbf{f}^T\mathbf{f}^+/\tau\right)}}{e^{\left(\mathbf{f}^T\mathbf{f}^+/\tau\right)} +
% 		\sum_i e^{\left(\mathbf{f}^T\mathbf{f}_i^-/\tau\right)}}\right] \\
%   &\geq \log\left(K\right)-I\left(\mathbf{f}, \mathbf{f}^+\right).
% \end{split}
% 	% \label{eq:bound}
% \end{equation}
\begin{align}
    	\displaystyle \mathop{\mathbb{E}}
	_{\substack{\left(\mathbf{z}, c, \mathbf{z}^+, c^+\right)\sim p_{\text{pos}}\\ 
			\left(\mathbf{z}_i,c^-_i\right)\stackrel{i.i.d.}{\sim} p\left(\mathbf{z}, \text{neg}\left(c\right)\right)}}
	&\left[-\log \frac{ e^{\left(\mathbf{f}^T\mathbf{f}^+/\tau\right)}}{e^{\left(\mathbf{f}^T\mathbf{f}^+/\tau\right)} +
		\sum_i e^{\left(\mathbf{f}^T\mathbf{f}_i^-/\tau\right)}}\right] \nonumber \\
  &\geq \log\left(K\right)-I\left(\mathbf{f}, \mathbf{f}^+\right). \label{eq:bound1}
\end{align}
\end{lemma}
\begin{proof}
% \noindent\emph{Proof}.  
Sample $\left(\mathbf{z}, c, \mathbf{z}^+, c^+\right)$ from $p_{pos}$ to generate images $G\left(\mathbf{z},c\right)$ and $G\left(\mathbf{z}^+, c^+\right)$, the representations $\mathbf{f}$ and $\mathbf{f}^+$ are encoded by encoder $E_c$.   
We then sample $K-1$ negative points $\left(\mathbf{z}_i,c_i^-\right)$ to generate images $G\left(\mathbf{z}_i,c_i^-\right)$, and finally obtain $K-1$ representations $\mathbf{f}^-_i, i=1,...,K-1$.
$\mathbf{f}^+$ and $\left\{\mathbf{f}_i^-\right\}$ can construct a new set $\left\{\mathbf{f}_k\right\}_{k=1}^K$, where the $k^{th}$ element is the representation generated from $p_{pos}$. 
The contrastive loss $\mathcal{L}_c$ is the categorical cross-entropy of classifying the positive sample correctly. The optimal probability for this loss is:

\begin{align}
    		p\left(d=k|\mathbf{f},\left\{\mathbf{f}_i\right\}_{i=1}^{K-1}\right)
		&= \frac{p\left(\mathbf{f}^+|\mathbf{f}\right) \prod_{l \neq k} p\left(\mathbf{f}^-_l\right)}
		{\sum_{j=1}^N p\left(\mathbf{f}^+|\mathbf{f}\right) \prod_{l \neq j}p\left(\mathbf{f}^-_l\right)}  \nonumber\\
		%---------------------------------------------------------------------------------------
		&=
		\frac{\frac{p\left(\mathbf{f}^+|\mathbf{f}\right)}{p\left(\mathbf{f}^+\right)}}{\frac{p\left(\mathbf{f}^+|\mathbf{f}\right)}{p\left(\mathbf{f}^+\right)}
			+\sum_{j \neq k}\frac{p\left(\mathbf{f}_j^-|\mathbf{f}\right)}{p\left(\mathbf{f}_j^-\right)},
		}\label{eq:prob}
\end{align}
where $[d=k]$ is the indicator that the $k^{th}$ sample is the 'positive' sample. Compare \Cref{eq:bound1} and \Cref{eq:prob}, we can obtain:
\begin{equation}
	e^{\left(\mathbf{f}^T\mathbf{f}^+/\tau\right)}	\propto \frac{p\left(\mathbf{f}^+|\mathbf{f}\right)}{p\left(\mathbf{f}^+\right)}.
\end{equation}
Inserting this back to the contrastive loss $\mathcal{L}_c$:
% \begin{equation}
% \small
% 	\begin{split}

% 	\end{split}.
% \end{equation}

\begin{align}
    \mathcal{L}^{\text{opt}}_{c} &=	\displaystyle \mathop{-\mathbb{E}}
    _{\substack{\left(\mathbf{z}, c, \mathbf{z}^+, c^+\right)\sim p_{\text{pos}}\\ \nonumber \left(\mathbf{z}_i,c^-_i\right)\stackrel{i.i.d.}{\sim} p\left(\mathbf{z}, \text{neg}\left(c\right)\right)}}
    \log \left[ \frac{\frac{p\left(\mathbf{f}^+|\mathbf{f}\right)}{p\left(\mathbf{f}^+\right)}}{\frac{p\left(\mathbf{f}^+|\mathbf{f}\right)}{p\left(\mathbf{f}^+\right)}
        +\sum_i\frac{p\left(\mathbf{f}_i^-|\mathbf{f}\right)}{p\left(\mathbf{f}_i^-\right)}
    }\right]\\
    %---------------------------------------------------------------------------------------
    &=\displaystyle \mathop{\mathbb{E}}
    _{\substack{\left(\mathbf{z}, c, \mathbf{z}^+, c^+\right)\sim p_{\text{pos}}\\ \nonumber \left(\mathbf{z}_i,c^-_i\right)\stackrel{i.i.d.}{\sim} p\left(\mathbf{z}, \text{neg}\left(c\right)\right)}}
    \log\left[1 + \frac{p\left(\mathbf{f}^+\right)}{p\left(\mathbf{f}^+|\mathbf{f}\right)}
    \sum_i\frac{p\left(\mathbf{f}_i^-|\mathbf{f}\right)}{p\left(\mathbf{f}_i^-\right)}
    \right]\\
\end{align}
\begin{align}
    %---------------------------------------------------------------------------------------
    &\approx  \displaystyle \mathop{\mathbb{E}}
    _{\substack{\left(\mathbf{z}, c, \mathbf{z}^+, c^+\right)\sim p_{\text{pos}}}}
    \log\left[1 + \frac{p\left(\mathbf{f}^+\right)}{p\left(\mathbf{f}^+|\mathbf{f}\right)}
    \left(K-1\right)\right. \nonumber\\
    &~~~~~~\left.\mathbb{E}_{\substack{\left(\mathbf{z}_i,c^-_i\right)\stackrel{i.i.d.}{\sim} p\left(\mathbf{z}, \text{neg}\left(c\right)\right)}}\frac{p\left(\mathbf{f}_i^-|\mathbf{f}\right)}{p\left(\mathbf{f}_i^-\right)}
    \right]\\
    &= \displaystyle \mathop{\mathbb{E}}
    _{\substack{\left(\mathbf{z}, c, \mathbf{z}^+, c^+\right)\sim p_{\text{pos}}}}
    \log\left[1 + \frac{p\left(\mathbf{f}^+\right)}{p\left(\mathbf{f}^+|\mathbf{f}\right)}
    \left(K-1\right)
    \right]\\
    %----------------------------------------------------------------------------------------
    &\geq \displaystyle \mathop{\mathbb{E}}
    _{\substack{\left(\mathbf{z}, c, \mathbf{z}^+, c^+\right)\sim p_{\text{pos}}}}
    \log\left[\frac{p\left(\mathbf{f}^+\right)}{p\left(\mathbf{f}^+|\mathbf{f}\right)}K\right]\\
    %----------------------------------------------------------------------------------------
    &=\log\left(K\right)-I\left(\mathbf{f},\mathbf{f}^+\right)
\end{align}
Minimizing the contrastive loss $\mathcal{L}_c\left(\mathbf{z},c\right)$ is equivalent to maximizing the mutual information between $\mathbf{f}$ and $\mathbf{f}^+$.
\end{proof}

\section{Implementation Details}
\label{sec:idetails}
% \subsection{Model setting}
% We employ Adam optimizer with $\beta_1=0$ and $\beta_2=0.99$ and learning rate $2e\text{-}4$. 
% ResNet18\cite{he2016deep} is backbone of $E_c$ and $E_\mathcal{Z}$. 
% Trade off is shown in Tab.~\ref{tab:tradeoff}. 
% Batch size is $128$ for $G$, $64$ for $D$, $265$ for $E_c$, and $32$ for $E_z$. 
% The dim of $\mathcal{L}_c$ is 128.
% Epoch is $350$ for CIFAR-10, $3,000$ for 3D-Chairs, $4,500$ for COIL-20.
% $L_{GAN}$ is hinge loss for CIFAR-10, and WGAN loss for other datasets.
% $z_n$ dim is 118, and $z_c$ dim is 10. 
% Follow Mukherjee \etal \cite{mukherjee2019clustergan}, we calculate the disentanglement accuracy of $D$ on MNIST, Fashion-MNIST and CIFAR-10 datasets.
% Follow Ojha \etal \cite{ojha2020elastic},  we calculate the disentanglement accuracy of $G$ on COIL-20, 3D-Chairs and 3D-Cars datasets.

% % Other setting can be found in Code. 
\texttt{CoDeGAN} comprises a generator $G$, a discriminator $D$, an encoder $E_c$, and an encoder $E_\mathbf{z}$. 
Except on the CIFAR-10 dataset, the encoders $E_c$ and $E_\mathbf{z}$ share weights except for the last layer.
% The experimental details, such as trade-offs, batch size, network architecture, and optimization details, are given in the following. 
% ``BN'' is the abbreviation for batch normalization, and ``SN'' is the abbreviation for spectral normalization~\cite{miyato2018spectral}.
The value of $\tau$ was all set to be 0.07.
The disentanglement accuracy, for MNIST, Fashion-MNIST, and CIFAR-10, is calculated following \cite{mukherjee2019clustergan}.
That for COIL-20, 3D-Chairs, and 3D-Cars is calculated following the work~\citep{ojha2020elastic}.

\subsection{MNIST}
On the MNIST dataset, the architecture  of \texttt{CoDeGAN} followed the work~\citep{mukherjee2019clustergan}, which was based on WGAN-GP~\citep{gulrajani2017improved}.
% The encoder's architecture is almost the same as the discriminator but with one more constitutional layer.
The batch size for the encoder, generator, and discriminator was 256, 64, and 64. 
The dimension of $\mathbf{z}$ was 30, and $c$ was encoded to be a one-hot vector.
The dimensions of the representations $\mathbf{f}$ and $\hat{\mathbf{z}}$ were set to be 120 and 30, respectively.
$\beta_1$ and $\beta_2$ were set to be 75 and 0.0005.
% LReLU activation with leak 0.2 was used. 
% Besides, Adam optimizer~\cite{kingma2014adam} with a learning rate of 0.0001 was adopted to train \texttt{CoDeGAN} for 400 epochs.
% Tab.~\ref{ArcMNIST} shows the network architecture of \texttt{CoDeGAN} on the MNIST dataset.

\subsection{Fashion-MNIST}
On the Fashion-MNIST dataset, the architecture  of \texttt{CoDeGAN} followed the work~\citep{mukherjee2019clustergan}, which was based on WGAN-GP~\citep{gulrajani2017improved}.
% The encoder's architecture is almost the same as the discriminator but with one more constitutional layer.
The batch size was set to be 64 for $G$ and $D$, 300 for $E$. 
The dimensions of $\mathbf{f}$, $\mathbf{z}$ and $\hat{\mathbf{z}}$ were 120, 40 and 40,  respectively. 
$\beta_1$ and $\beta_2$ were set to be 100 and 0.0005.
% LReLU activation with leak 0.2 was used. 
% Besides, Adam optimizer~\cite{kingma2014adam} with a learning rate of 0.0001 was adopted to train \texttt{CoDeGAN} for 400 epochs.
% In the few labels setting, $0.17\%$ of the training samples have labels.
% \texttt{CoDeGAN} was trained for 600 epochs except in few label setting with contrastive pretraining, in which \texttt{CoDeGAN} was trained for 300 epochs.\par

In pretraining,  contrastive pretraining and meta pretraining are used to train $E_c$.
Besides, the parameters of $E_c$ were fixed in the first 200 epochs when training \texttt{CoDeGAN}. 
Some details  are listed in the following:
\begin{itemize}%[leftmargin=*]
	\item [1] 
	In contrastive pretraining, SimCLR~\citep{hinton2020SIMCLR} was used to pretrain the encoder $E_c$ for 200 epochs.
	% We augmented each sample to produce positive pairs, and left samples and their augmentations are used to produce negative pairs.
	% The encoder $E_c$ was fine-tuned for 20 epochs using 100 labeled images in the few label setting.       
% % 	SimSiam~\cite{chen2021Siamese} was used pretrain the encoder $E_C$ for ..... \textcolor{red}{add}
% % 	NNCLR~\cite{ermolov2021whitening} was used pretrain the encoder $E_c$ for ... \textcolor{red}{add}
	\item [2]
	In meta pretraining, MAML~\citep{finn2017model} was used to pretrain the encoder $E_c$ for 80 epochs, where supervised contrastive loss~\citep{2020Supervised} was used. 
% 	The MNIST dataset was chosen as the training set.
	The number of tasks was set to 10000, and each task included 5 ways.
	Both the support set and query set contained 20 shots, and 4 tasks were used in an iteration. 
% 	The meta step size was 0.0001, and the step size was 0.01.
	% The encoder $E_c$ was fine-tuned for 300 epochs with 100 labeled images in the few labels experiment. 
\end{itemize}

\subsection{CIFAR-10}
On the CIFAR-10 dataset, the network architecture of \texttt{CoDeGAN} followed SNGAN~\citep{miyato2018spectral}, where the encoder's architecture was ResNet18~\citep{he2016deep}.
The batch size for $G$, $D$, $E_c$, and $E_\mathbf{z}$ were 128, 64, 256, 32.
The dimensions of $\mathbf{f}$, $\mathbf{z}$, and $\hat{\mathbf{z}}$ were set to be 128, 118, and 118, and $c$ was encoded to be a 10-dimensional one-hot vector. 
% In the few labels experiments, $1\%$ of the training samples were with labels.
% Adam optimizer~\cite{kingma2014adam} was adopted with a learning rate of 0.0002 in all experiments. 
% The network architecture and the ResBlock are shown in Tab.~\ref{ArcCIFAR-10} and Tab.~\ref{table_ResBlock}. 
% $\beta_1$, $\beta_2$, and training epoch numbers were set differently in variant \texttt{CoDeGAN}s as in Tab.~\ref{paraCIFAR}.
In pretraining, contrastive pretraining and meta pretraining are used to train $E_c$. 
Besides, the parameters of $E_c$ were fixed in the first 100 epochs when training \texttt{CoDeGAN}. 
Some details in pretraining are listed in the following:
\begin{itemize} %[leftmargin=*]
	\item [1] 
	In contrastive pretraining, SimCLR~\citep{hinton2020SIMCLR} was used to train $E_c$ for 200 epochs. 
	% We augmented each sample to produce positive pair and left samples, and their augmentations are used to produce negative pairs.
	% The encoder $E_c$ was fine-tuned for 20 epochs using 500 labeled images in the few label setting.   
% 	SimSiam~\cite{chen2021Siamese} was used pretrain the encoder $E_C$ for ..... \textcolor{red}{add}
% 	NNCLR~\cite{ermolov2021whitening} was used pretrain the encoder $E_c$ for ... \textcolor{red}{add}
	
	\item [2]
	In meta pretraining, MAML~\citep{finn2017model} was used to pretrain the encoder $E_c$ for 40 epochs, where supervised contrastive loss~\citep{2020Supervised} was used. 
	The Tiny-ImageNet dataset was chosen as the training set.
	The number of tasks was set to 10000, and each task included 10 ways.
	The  support set contained 10 shots, the query set contained 20 shots, and 4 tasks were used in an iteration. 
	% The meta step size was 0.0001, and the step size was 0.01.
	% The encoder $E_c$ was fine-tuned for 300 epochs with 500 labeled images in the few labels experiment. 
\end{itemize}

\subsection{COIL-20}
On the COIL-20 dataset, the network architecture of \texttt{CoDeGAN} followed WGAN-GP\citep{gulrajani2017improved} with BCE Loss.
% % and the encoder’s architecture was ResNet18~\cite{he2016deep}
% % \textcolor{red}{followed the work~\cite{mukherjee2019clustergan}, which was based on DCGAN~\cite{goodfellow2014generative}.}
% The encoder's architecture is ResNet18 with two MLP layers.
The batch size for $G$, $D$, $E_c$ and $E_\mathbf{z}$ were 256, 128, 512, 64. 
The dimension of $\mathbf{z}$ was 118, and $c$ was encoded to be a 20-dimensional one-hot vector.
The dimensions of the representations $\mathbf{f}$ and $\hat{\mathbf{z}}$ were set to be 128 and 118, respectively.
$\beta_1$ and $\beta_2$ were set to be 150 and 0.0005.
% Besides, Adam optimizer~\cite{kingma2014adam} with a learning rate of 0.0002 and betas $(0.5,0.999)$  was adopted to train \texttt{CoDeGAN} for 4500 epochs.
% Tab.~\ref{ArcCOIL} and \ref{table_ResBlock_other} show the network architecture of \texttt{CoDeGAN} for COIL-20.

In contrastive pretraining, SimCLR~\citep{hinton2020SIMCLR}, SimSiam\citep{chen2021Siamese}  and NNCLR~\citep{ermolov2021whitening} were used to train $E_c$. 
Besides, the parameters of $E_c$ were fixed in the first 1,500 epochs when training \texttt{CoDeGAN}. 
Some details in pretraining are listed in the following:
({\romannumeral1}) SimCLR was used to pretrain the encoder $E_c$ for 76 epochs. 
	% We augmented each sample to produce positive pair, and left samples, and their augmentations are used to produce negative pairs.
	Adam optimizer was adopted with a learning rate of 0.0002 and betas $(0.5,0.99)$, $\beta_1$ was 100. 
	The batch size is 256.
({\romannumeral2}) SimSiam was used to pretrain the encoder $E_c$ for 1,585 epochs, $\beta_1$ was 50.
	The other settings were the same as the SimCLR method.
({\romannumeral3}) NNCLR was used to pretrain the encoder $E_c$ for 209 epochs, $\beta_1$ was 50. 
	The other settings were the same as the SimCLR method.

\subsection{3D-Chairs and 3D-Cars}
On the 3D-Chairs and 3D-Cars datasets, the network architecture of \texttt{CoDeGAN} followed WGAN-GP~\citep{gulrajani2017improved} with BCE Loss.
% % Some differences are list in the following.
% % the architecture of \texttt{CoDeGAN} \textcolor{red}{followed the work~\cite{mukherjee2019clustergan}, which was based on WGAN-GP~\cite{gulrajani2017improved}.}
% The encoder's architecture was ResNet18 with two MLP layers.
% The batch size for $G$, $D$, $E_c$, and $E_\mathbf{z}$ were 48, 60, 80, 40. 
The dimension of $\mathbf{z}$ was 118. and $c$ was encoded to be a 10-dimensional one-hot vector.
The dimensions of the representations $\mathbf{f}$ and $\hat{\mathbf{z}}$ were set be 128 and 118, respectively.
$\beta_1$ and $\beta_2$ were set to be 100 and 0.0005.
% Besides, Adam optimizer~\cite{kingma2014adam} with a learning rate of 0.0002 and betas $(0.5,0.999)$ was adopted to train \texttt{CoDeGAN} for 3000 epochs.
% Tab.~\ref{Arc3D} and \ref{table_ResBlock_other} show the network architecture of \texttt{CoDeGAN} on the 3D-Chairs and 3D-Cars datasets.

In contrastive pretraining, for 3D-Chairs, the parameters of $E_c$ were fixed in the first 1,000 epochs when training \texttt{CoDeGAN}, for 3D-Cars, the fixed epoch number and $\beta_1$ were 200 and 80.  
Some details in pretraining are listed in the following:
({\romannumeral1}) was used to pretrain the encoder $E_c$ for 122 epochs. 
% We augmented each sample to produce positive pairs, and left samples, and their augmentations are used to produce negative pairs.
Adam optimizer was adopted with a learning rate of 0.0002 and betas $(0.5,0.99)$, $\beta_1$ was 75. 
The batch size is 256.
({\romannumeral2}) SimSiam was used pretrain the encoder $E_c$ for 989 epochs.
The other settings were the same as the SimCLR method.
({\romannumeral3}) NNCLR was used pretrain the encoder $E_c$ for 405 epochs, $\beta_1$ was 50. 
The other settings were the same as the SimCLR method.

% \begin{table}[h]
% % 	\small
% 	\centering\caption{The Architectures of ResBlock on the 3D-Chairs and 3D-Cars datasets.}
% 	\centering{
% 		\resizebox{0.95\textwidth}{!}{
% 			\begin{tabular}{|l|l|l|}
% 				\hline
% 				\textbf{GenBlock} & \textbf{DisBlock1} & \textbf{DisBlock2} \\ \hline 
% 				upsample with nearest interpolation&3$\times$3 conv. stride 1. RELU& 3$\times$3 conv. stride 1. RELU\\ \hline
% 				3$\times$3 conv. stride 1. BN. RELU&3$\times$3 conv. stride 1 &3$\times$3 conv. stride 1. RELU\\ \hline
% 				3$\times$3 conv. stride 1. BN. RELU&3$\times$3 pooling. stride 1. RELU & \\ \hline
% 			\end{tabular}		
% 	}}
% 	\label{table_ResBlock_3D}
% \end{table}
% \newpage
\section{More Results}

\subsection{Robustness to Intra-class Variations} 
\begin{figure}[ht]
	\centering
	\subfloat[Square]{%
		\includegraphics[width=0.4\linewidth]{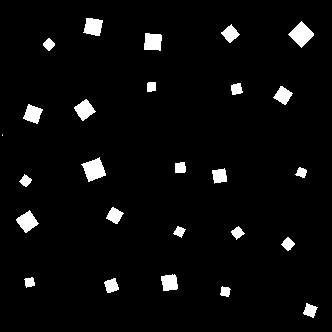}%
		\label{fig:dsprites_a}%
	}\hspace{0.25cm}
	\subfloat[Ellipse]{%
		\includegraphics[width=0.4\linewidth]{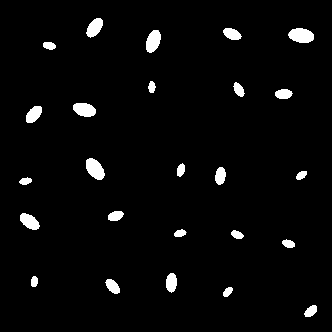}%
		\label{fig:dsprites_b}%
	}\hspace{0.25cm}
	\subfloat[Heart]{%
		\includegraphics[width=0.4\linewidth]{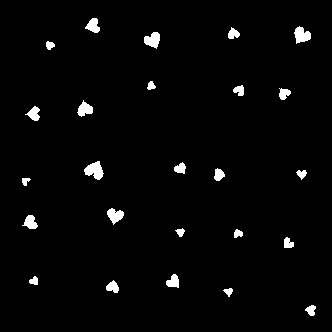}%
		\label{fig:dsprites_c}%
	}\hspace{0.25cm}
	\subfloat[ACC, NMI, and ARI]{%
		\includegraphics[width=0.42\linewidth]{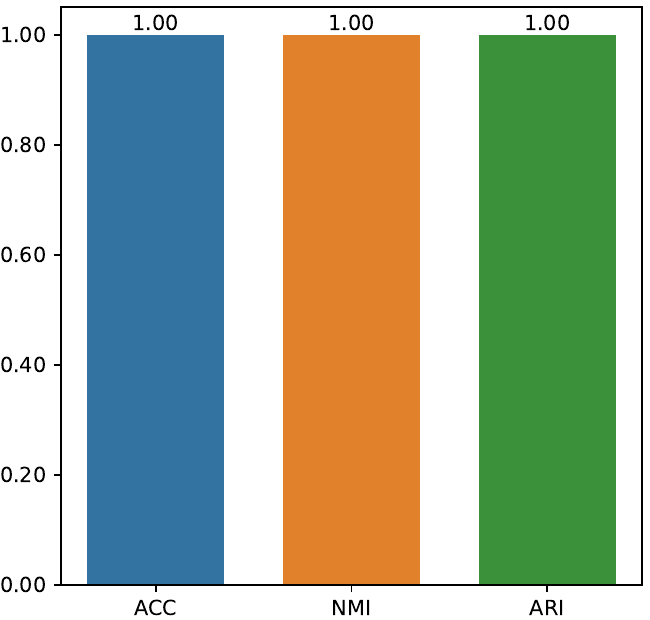}%
		\label{fig:dsprites_d}%
	}
	\caption{\textbf{Images generated by \texttt{CoDeGAN} trained on dSprites.} (a), (b), and (c) show the images corresponding to the three shapes, respectively. \texttt{CoDeGAN} is robust to intra-class changes, such as scale, rotation, and translation.}
	\label{fig:dsprites}
\end{figure}

To test the robustness of \texttt{CoDeGAN} to intra-class variations, we conduct experiments on the dSprites dataset~\citep{dsprites17}.
The dSprites dataset consists of 737,280 binary $64\times64$ images of 2D shapes.
On the dSprites dataset, the network architecture of CoDeGAN followed SNGAN~\citep{lin2020infogan}, and the encoder's architecture followed the Q-network of InfoGAN-CR~\citep{lin2020infogan}.
The GAN's loss was the traditional JSD loss as in InfoGAN-CR~\citep{lin2020infogan}. 
The batch size was set to 150 for the generator, 300 for the discriminator, and the encoder. 
The dimensions of $\mathbf{f}$, $\mathbf{z}$, and $\hat{\mathbf{z}}$ were 40, 52 and 52.
In addition, $\beta_1=1$, and $\beta_2=0.0001$.
% LReLU activation with leak 0.1 was used. 
% Adam optimizer~\cite{kingma2014adam} was used with learning rate 0.001 for $G$ and $E$, and 0.002 for $D$.
$E_c$ was pretrained by SimCLR for 100 epochs and further fine-tuned for 20 epochs using few labeled images.
The parameters of $E_c$ were fixed in the first 100 epochs, and CoDeGAN was trained for 300 epochs.
727,280 images were used for training, and the rest 10,000 for testing.
$0.21\%$ of the training samples were set to have labels.

It has five independent latent factors: shape, scale, rotation, $x$, and $y$ positions of a sprite.
The shape contains three categories: square, ellipse, and heart.
\Cref{fig:dsprites} shows the disentangled results on dSprites dataset.
\Cref{fig:dsprites_a,fig:dsprites_b,fig:dsprites_c} show the generation results of changing factor $c$ and $\mathbf{z}$, where the shape ``square'', `` ellipse'', and ``heart'' are generated orderly.
These results reveal our \texttt{CoDeGAN} with few labels is robust to intra-class changes, such as scale, rotation, and translation.
This should own to our loss function, which, coupled with contrastive prior, can utilize few labels to the largest extent.
The ACC, NMI, and ARI are all close to 1, even though only 0.21\% of labels are involved.

\subsection{Sensitivity to Tiny Inter-Class Variation}

\begin{figure}[ht]
	\centering
	\subfloat{%
		\includegraphics[width=0.9\linewidth]{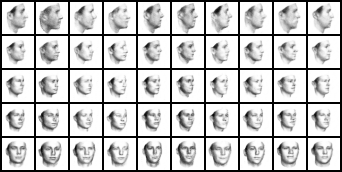}%
	}\\
	\subfloat{%
		\includegraphics[width=0.9\linewidth]{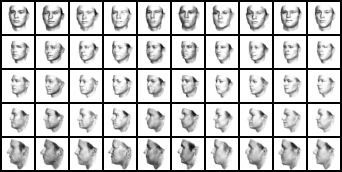}%
	}
	\caption{\textbf{Images generated by \texttt{CoDeGAN} trained on Faces.} Each row corresponds to one latent factor value. Experiments show \texttt{CoDeGAN} has the potential to disentangle class with a tiny variation.}
	\label{fig:faces}
\end{figure}

\begin{figure*}[ht]
	\centering
	\subfloat{\includegraphics[width=0.3\linewidth]{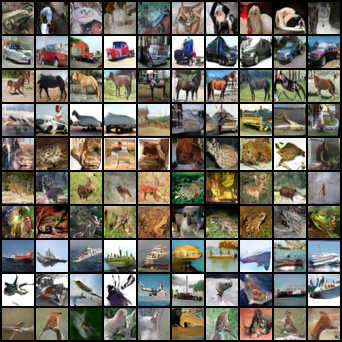}}\hspace{0.1cm}
	\subfloat{\includegraphics[width=0.3\linewidth]{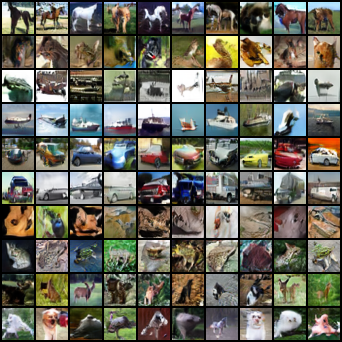}}\hspace{0.1cm}
	\subfloat{\includegraphics[width=0.3\linewidth]{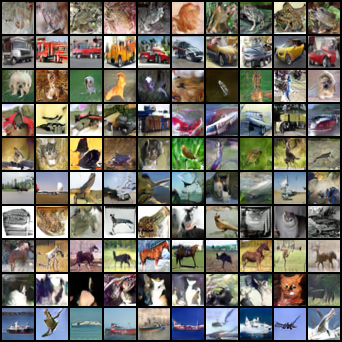}}\\
	\subfloat{\includegraphics[width=0.3\linewidth]{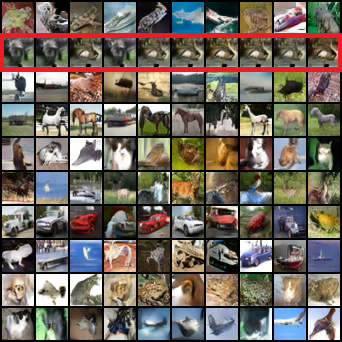}}\hspace{0.1cm}
	\subfloat{\includegraphics[width=0.3\linewidth]{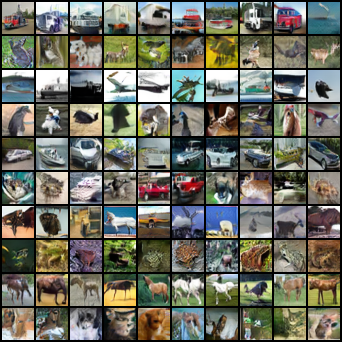}}
	\caption{\textbf{5 Times Generation Results for CIFAR-10 with $L_c$ on the layers $L_2+L_3+L_4$.} Red box highlights mode collapse/dropping.}
	\label{fig:cifar10_L2}
\end{figure*}
\begin{figure*}[ht]
	\centering
	\subfloat{\includegraphics[width=0.3\linewidth]{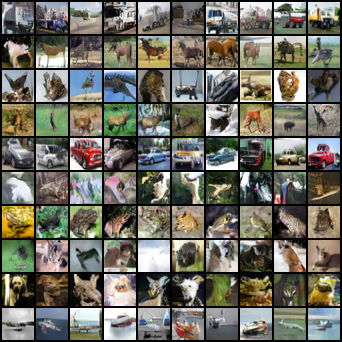}}\hspace{0.1cm}
	\subfloat{\includegraphics[width=0.3\linewidth]{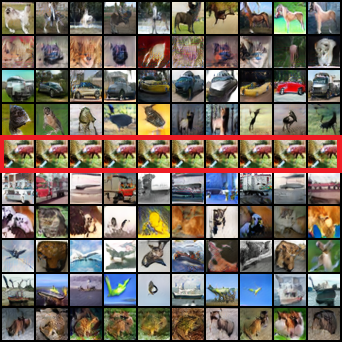}}\hspace{0.1cm}
	\subfloat{\includegraphics[width=0.3\linewidth]{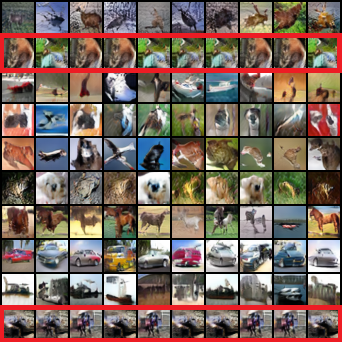}}\\
	\subfloat{\includegraphics[width=0.3\linewidth]{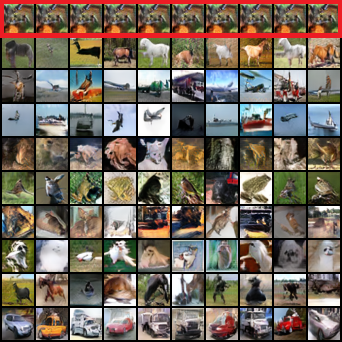}}\hspace{0.1cm}
	\subfloat{\includegraphics[width=0.3\linewidth]{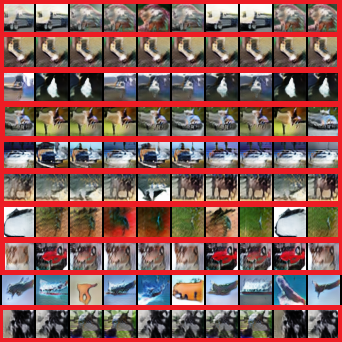}}
	\caption{\textbf{5 Times Generation Results for CIFAR-10 with $L_c$ on all layers $L_1+L_2+L_3+L_4$.} Red box highlights mode collapse/dropping.}
	\label{fig:cifar10_L1}
\end{figure*}

To test the sensitivity of \texttt{CoDeGAN} to tiny inter-class variation, we conduct experiments on the Faces datasets~\citep{faces}.
The Faces dataset contains variation factors such as azimuth (pose), elevation, and lighting.
On the Faces dataset, the network architecture of CoDeGAN followed InfoGAN~\citep{chen2016infogan},  but without batch normalization in the discriminator.
The encoder's architecture was the same as the discriminator except for the last layer.
The GAN's loss followed the WGAN-GP~\citep{gulrajani2017improved}.
The batch size for the generator and the discriminator was set to 300, and that for the encoder was 1500. 
The dimension of $\mathbf{f}$, $\mathbf{z}$ and $\hat{\mathbf{z}}$ was 118.
Besides, $\beta_1=10$, and $\beta_2=10$.
% LReLU activation with leak 0.2 was used. 
% Adam optimizer~\cite{kingma2014adam} was adopted with a learning rate of 0.0005 for $G$ and $E$, 0.0002 for $D$.
We trained CoDeGAN for 600 epochs on the Faces dataset. 
$1\%$ of the training samples were with labels.
$E_c$ was pretrained by SimCLR for 200 epochs and further fine-tuned for 20 epochs using few labeled images.
When training CoDeGAN, the parameters of $E_c$ were fixed in the first 200 epochs.

To build a training set, 10,000  face prototypes were synthesized, and ten face images were generated for each face prototype with a pose from $-90^\circ$ to $+90^\circ$.
\Cref{fig:faces} shows the disentangled results on the Faces dataset, where each row corresponds to a different head pose. 
We discretized the continuous pose change to discrete pose change; the interval between each group is $20^{\circ}$.
We observe in \Cref{fig:faces}, compared with digits, objects, or shapes, the images in adjacent categories are very similar.
This demonstrates \texttt{CoDeGAN} also has the potential to disentangle class with a tiny variation.

\subsection{Ablation Study} 
\label{sec:appendix_abl}

We study the effects of contrastive loss on different feature layers.
The features from four layers of the encoder network $E_c$ at different resolutions $\left(L_1=32^2, L_2=16^2, L_3=8^2, L_4=4^2 \right)$ were extracted. 
Here, $L_4$ denotes the highest feature layer, and $L_1$ denotes the lowest feature layer, which is most \emph{close} to the generated image.
The features were down-sampled four times as well as flattened.  
Then, the multi-scale features were fed to a two-layer MLP and reduced to the same 128 dimensions.

We show generated images with $\mathcal{L}_c$ on different feature layers for CIFAR-10, as shown in 
%Fig.~\ref{fig:cifar10_L4}, Fig.~\ref{fig:cifar10_L3}, 
\Cref{fig:cifar10_L2,fig:cifar10_L1}.
% show the generation results with $\mathcal{L}_c$ on $L_4$, $L_3+L_4$, $L_2+L_3+L_4$ and $L_1+L_2+L_3+L_4$.
% For 3D-Chairs,Fig.~\ref{fig:chairs_L4}, Fig.~\ref{fig:chairs_L3}, Fig.~\ref{fig:chairs_L2} and Fig.~\ref{fig:chairs_L1} show the generation results with $\mathcal{L}_c$ on $L_4$, $L_3+L_4$, $L_2+L_3+L_4$and $L_1+L_2+L_3+L_4$. 
Mode collapse/dropping sometimes occurs when adding contrastive loss on lower feature layers.
% The generated images with mode collapse/dropping can be found in Fig.~\ref{fig:cifar10_L2}, \ref{fig:cifar10_L1}, \ref{fig:chairs_L2} and \ref{fig:chairs_L1}. 
This tendency may be explained by the fact that the similarity constraints relaxing on low-layer features are minimal.
This is because these layers are close to the image, and the model capacity of the low-layer sub-networks of encoder $E_c$ is restricted.

\bibliographystyle{elsarticle-harv} 
\bibliography{egbib}





% \end{thebibliography}
\end{document}